\documentclass[10pt,twocolumn,letterpaper]{article}
\usepackage[pagenumbers]{cvpr}
\usepackage{amsmath,amssymb,booktabs,graphicx,multirow,tabularx,enumitem}
\usepackage{microtype}
\definecolor{cvprblue}{rgb}{0.21,0.49,0.74}
\usepackage[pagebackref,breaklinks,colorlinks,allcolors=cvprblue]{hyperref}
\hypersetup{pdftitle={Does Continual Anomaly Detection Require Continual Training?},pdfauthor={JOONGWON CHAE, Lihui Luo, Yang Liu, Yiyi Cai, Roger Guangxuan Liu, Wenjie Li, Peiwu Qin, Runming Wang}}
\def\confName{CVPR}
\def\confYear{2027}
\newcommand{\Iauc}{I-AUROC}
\newcommand{\Pauc}{P-AUROC}
\newcommand{\Pap}{P-AP}

\newcommand{\suppref}[1]{Appendix~\ref{#1}}
\title{Does Continual Anomaly Detection Require Continual Training?}
\author{\parbox{\dimexpr\textwidth-2\tabcolsep\relax}{\centering
\normalsize JOONGWON CHAE$^{1}$\quad Lihui Luo$^{1}$\quad Yang Liu$^{1}$\quad Yiyi Cai$^{1}$\\[3pt]
\normalsize Roger Guangxuan Liu$^{1}$\quad Wenjie Li$^{2}$\quad Peiwu Qin$^{3}$\quad Runming Wang$^{1}$\\[7pt]
\small $^{1}$Tsinghua University\\
\small $^{2}$Shanghai Innovation Institute\\
\small $^{3}$Chinese Medicine Guangdong Laboratory
}}

\usepackage{placeins}

\begin{document}
\maketitle
\begin{abstract}
Continual anomaly detection learns the normal appearance of new products while continuing to inspect earlier ones. This setting arises when product categories arrive over time and historical training images are no longer available. Existing methods often update a shared detector and constrain these updates or replay historical information to reduce forgetting. We ask whether adding a new category requires updating detectors already built for earlier categories. To answer this question, we compare six detector families on MVTec AD and VisA using baselines that build each category's detector once, keep it fixed, and select it for an unlabeled test image. Geometry-Consistent Routing (GCR) makes this selection by comparing test patches with each category's normal references. These comparisons distinguish detection lost through routing from changes due to learning, and initial gains from those retained after later tasks. Preserving and routing detectors extends inspection across both datasets without retraining earlier detectors; routing also remains correct for all 10,900 images in a mixed industrial and medical pool of 30 categories. Further learning helps in some settings and, with real historical images, can replace separate detectors with a stronger shared model. However, initial gains can disappear during later tasks, while low forgetting can accompany weak detection throughout. Adding categories and benefiting from further detector updates are therefore separate questions.
\end{abstract}

\section{Introduction}
In manufacturing inspection, new products arrive over time while earlier products remain in production. The system must therefore learn the normal appearance of each new product and continue to detect defects in products it has already seen. When historical training images are unavailable, it cannot retrain on all products together. Continual anomaly detection (CAD) studies this sequential access to normal training images~\cite{dne,ucad,iuf,cdad,replaycad}.

Many CAD methods learn new categories by updating a shared detector while trying to retain its earlier detection ability. Since the same weights serve several categories, an update for a new product can also change predictions for earlier products. Existing methods limit this change by constraining weight updates~\cite{iuf,cdad} or reusing historical information through stored feature distributions~\cite{dne} and generated images~\cite{replaycad}.

\textbf{Why must learning a new product also change the detectors for earlier products?} If a detector still inspects its product reliably, the system can keep it fixed and add a detector for the new category. UCAD, for example, stores category-specific prompts and normal memory banks on a frozen encoder~\cite{ucad}. Adding a category and updating earlier detectors are thus separate design choices.

We study these choices using two baselines that leave earlier detectors unchanged. The first stores normal patch features from a frozen pretrained encoder; the second trains an independent detector once per category. As categories arrive, the system adds their references or detectors without changing earlier weights or normal references. At test time, it must select a detector without knowing the image's category. We propose \textbf{Geometry-Consistent Routing (GCR)} for this step, selecting the category whose normal bank best matches the patches in the test image.

This choice matters because normal cable structure can receive high anomaly scores when compared with metal-nut references. To measure the effect of selection, we hold the features, memory banks, and scoring rule fixed and compare each router with an oracle that supplies the true category. The resulting difference shows how routing changes the performance of the same stored detectors.

\begin{figure*}[t]
\centering
\includegraphics[width=\textwidth]{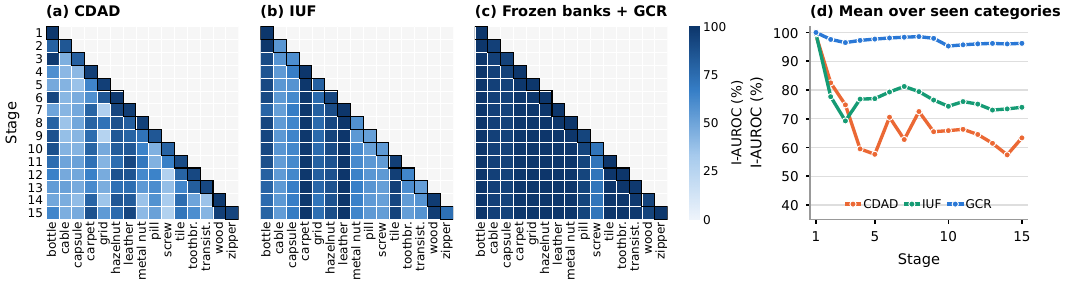}
\caption{\textbf{A small forgetting score does not necessarily indicate a strong detector.} Image AUROC across a 15-task MVTec AD stream, with one category per task. Each column follows a category from its first learning step (outlined diagonal) to the final task (bottom row). CDAD loses accuracy on earlier categories, whereas IUF changes less but starts with lower accuracy on several categories. Preserved banks with GCR keep each category's predictions unchanged. Stage means include all categories seen so far, so their composition changes as categories arrive.}
\label{fig:stream}
\end{figure*}

The preserved detectors also provide a reference for what further learning adds. We examine whether learning improves detection within a category, helps compare anomaly scores across categories, or allows one shared model to replace separate detectors. We then follow each category from acquisition through later tasks to see whether an initial gain survives. This distinction matters because final accuracy reflects both initial learning and later change, whereas forgetting measures do not reveal how well the detector worked at the start (Figure~\ref{fig:stream}).

Across six detector families on MVTec AD and VisA, preserved detectors with GCR cover the evaluated category streams without retraining earlier detectors. Learning improves some comparisons, but later updates can erase an initial gain, and weak detection can coexist with low forgetting. On MVTec, joint training with real historical images replaces separate models with a stronger shared detector; sequential generated replay falls below both. We also expand routing to \emph{Mixed30}, a pool of 30 industrial and medical categories, where GCR selects the true category for every test image.

Preservation also has a cost: separate neural detectors require more storage as categories arrive, while shrinking patch banks removes references used for localization. We therefore measure storage, acquisition time, and inference latency alongside detection accuracy. Our contributions are:
\begin{itemize}[leftmargin=*,nosep]
\item We propose GCR to select among preserved memory banks or per-category detectors. These baselines establish the detection available after learning each category once, providing a reference for the gains and costs of further training.
\item By tracking acquisition and later performance, we show when initial gains survive, when they disappear, and how low forgetting can accompany weak detection. Controls on routing, stored statistics, and bank size identify changes that remain possible even with fixed detector weights.
\end{itemize}

\section{Related Work}
\paragraph{Representing normal appearance.}
Anomaly detectors compare test images with normal examples or models learned from them, but differ in where they store normal information. Reference-based methods retain normal patch features~\cite{patchcore,pni} or local Gaussian statistics~\cite{padim}, while normalizing flows learn feature densities~\cite{differnet,cflow,fastflow}. Other methods encode normal appearance in network weights through reconstruction~\cite{ssimae,draem}, compact feature objectives~\cite{patchsvdd,vaesvdd,cfa}, teacher--student distillation~\cite{stfpm,rd,rdnormal}, or synthetic anomaly classification~\cite{anoseg,simplenet}. Diffusion models reconstruct normal images~\cite{anoddpm,ddad,diffusionad} or generate anomalies~\cite{anomalygen}, while unified models share components across categories~\cite{uniad,onenormal,ditad,uniref,dinomaly}.

\paragraph{Continual learning with a shared detector.}
Shared-model CAD methods update the detector across tasks while trying to retain earlier performance. DNE adapts its encoder and stores historical feature distributions~\cite{dne}; IUF constrains reconstruction learning~\cite{iuf}, CDAD projects gradients away from earlier feature subspaces~\cite{cdad}, and ReplayCAD supplies synthetic historical images from a learned generator~\cite{replaycad}. Pixel-level continual benchmarks~\cite{cadbenchmark} and lifelong adaptation~\cite{lifelongad} also study sequential data access and changes in normal appearance.

\paragraph{Preserving and routing category-specific models.}
Another approach keeps task-specific models and selects among them at test time, as in Expert Gate~\cite{expertgate}. In anomaly detection, UCAD combines a frozen backbone with category prompts and patch memory banks~\cite{ucad}, while Dynamic Expert Routing~\cite{der} and MECAD~\cite{mecad} route inputs across category-specific modules.

\paragraph{Pretrained representations and continual adaptation.}
Vision--language pretraining supports zero- and few-shot detectors such as WinCLIP, AnomalyCLIP, PromptAD, and AnomalyGPT~\cite{laion,winclip,anomalyclip,promptad,anomalygpt}. In continual classification, GDumb~\cite{gdumb}, SimpleCIL and APER~\cite{simplecil}, and First Session Adaptation~\cite{fsa} study stored data and frozen or adapted representations as alternatives to repeated weight updates.

\section{Method and Study Design}
\label{sec:design}
\subsection{Problem Setup and Preserved Baselines}
Normal training images $D_t$ arrive at task steps $t=1,\dots,T$. Each step introduces one or more new product categories into the observed set $\mathcal C_t$, and $\tau(c)$ denotes the step at which category $c$ first appears. Training images have category labels, whereas test images arrive without labels from any category in $\mathcal C_t$.

\begin{figure*}[t]
\centering
\includegraphics[width=\textwidth]{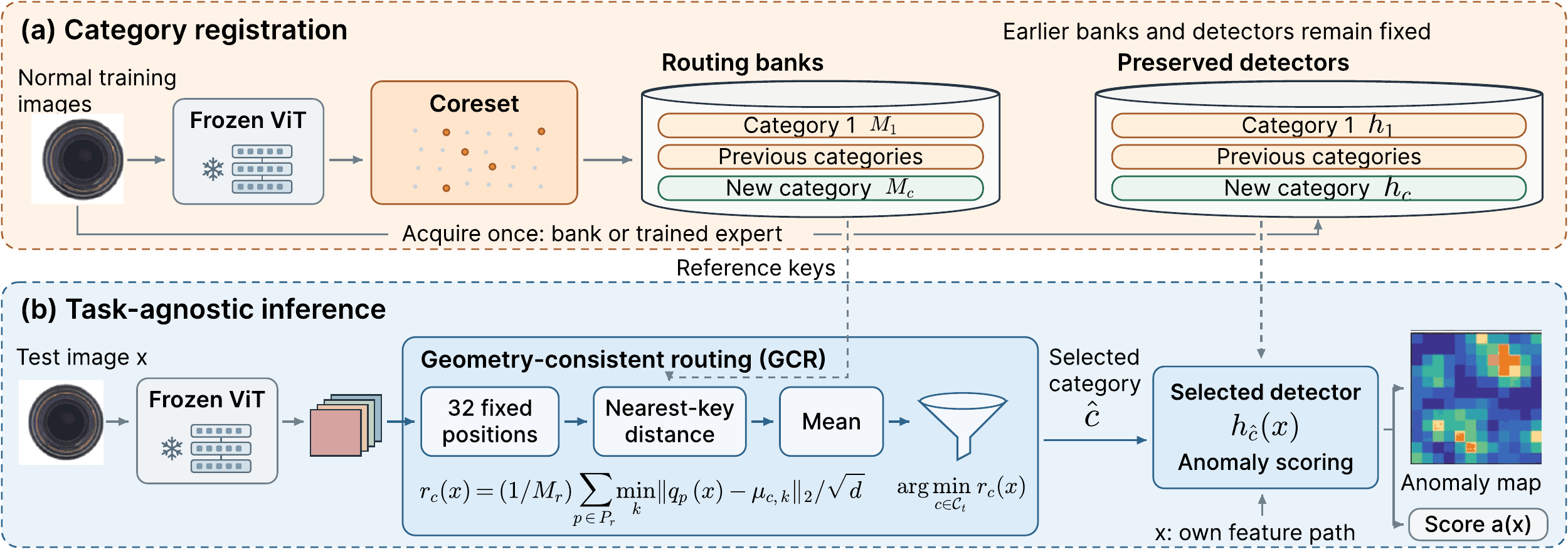}
\caption{\textbf{Preserving per-category detectors and routing test images with GCR.} Normal images from a new category provide its routing bank $M_c$ and detector $h_c$, while earlier banks and detectors remain fixed. At inference, GCR compares sampled test patches with the stored banks, selects a category, and runs its detector. Feature clouds and output maps are illustrative.}
\label{fig:gcr}
\end{figure*}

Our preserved baseline builds a detector $h_c$ once, when category $c$ arrives, and then keeps its parameters and normal references fixed. Given an unlabeled test image $x$ at step $t$, a router $R_t$ selects a category $\hat c_t(x)\in\mathcal C_t$. The selected detector returns an image anomaly score $a_t(x)$ and a pixel anomaly map $m_t(x)$:
\begin{equation}
 \hat c_t(x)=R_t(x),\qquad (a_t(x),m_t(x))=h_{\hat c_t(x)}(x).
 \label{eq:routing}
\end{equation}
To evaluate each detector without routing errors, we also use \emph{oracle selection}, which supplies the true test category ($\hat c_t(x)=c$). Adding a category expands the candidate set $\mathcal C_t$ but leaves earlier detectors unchanged, so each detector's oracle performance stays constant across the stream (Figure~\ref{fig:gcr}). Routed performance additionally depends on which detector is selected from the growing candidate set.

We use two forms of preserved detector. A \emph{patch memory bank} stores normal features from a frozen pretrained encoder and scores each test patch by its nearest-reference distance. A \emph{preserved neural detector} is an independent Dinomaly model trained once on category $c$ and kept fixed thereafter. In both cases, registration uses only the incoming normal images $D_t$, and inference runs the selected detector.

\subsection{GCR: Selecting the Appropriate Detector}
\label{sec:selector}
To route an image $x$, we compare its features with a normal reference bank $M_c=\{\mu_{c,k}\}_{k=1}^{K}$ for each candidate category $c\in\mathcal C_t$. The bank collects patches from many training images and therefore contains several views and normal variations. A single test image, however, shows one instance and need not contain every pattern in the bank.

UCAD's selector~\cite{ucad} searches in the \emph{bank-to-image} direction: for each stored reference $\mu_{c,k}$, it finds the nearest test patch and sums these distances over $k$. It therefore measures how much of the bank appears in one image, which can penalize a category whose bank contains normal patterns absent from that image.

GCR instead searches in the \emph{image-to-bank} direction, as in patch anomaly scoring. It asks whether the patches present in $x$ have close normal references in $M_c$. For $d$-dimensional query features $q_p(x)$ at $M_r=32$ fixed spatial positions $P_r$, GCR averages the nearest-reference distances and selects the category with the smallest mean:
\begin{equation}
 \begin{aligned}
 r_c(x)&=\frac{1}{M_r}\sum_{p\in P_r}\min_k
 \frac{\|q_p(x)-\mu_{c,k}\|_2}{\sqrt d},\\
 \hat c_t(x)&=\arg\min_{c\in\mathcal C_t}r_c(x).
 \end{aligned}
 \label{eq:gcr}
\end{equation}
We compare GCR with UCAD's selector and a global \emph{CLS baseline}. The latter uses cosine similarity between the test image's normalized class-token feature and each category's class-token centroid. Across selectors, we hold the encoder, reference banks, and patch-scoring rule fixed.

\subsection{Initial Gains, Retention, and Routing}
\label{sec:gaussian}
\label{sec:decomposition}
A higher score when a category is first learned need not remain higher after later tasks. To track this change, we compare continual adaptation with preserved versions of DNE~\cite{dne}, using the same categories and task orders (Table~\ref{tab:controls}). The comparison distinguishes the initial difference from preservation, the change during later tasks, and the effect of label-free selection in the preserved baseline.

DNE models normalized image features with Gaussians and scores anomalies by Mahalanobis distance. \emph{Adapted DNE} ($\mathrm{adapt}$) updates the encoder and Gaussian statistics sequentially. Its preserved counterpart fixes the initial encoder and stores one Gaussian per category. We evaluate these stored Gaussians with \emph{Separate Oracle} ($\mathrm{oracle}$), which uses the true category, and \emph{Separate Min} ($\min$), which takes the minimum distance across all stored Gaussians without category labels.

Let $A_{s,c}$ denote detection performance on category $c$ after step $s$, with $T$ the final task. We define the initial acquisition difference $G_c$ relative to the preserved oracle, later change $D_c$, and label-free selection difference $Q_c$ as
\begin{equation}
\begin{aligned}
 G_c&=A^{\mathrm{adapt}}_{\tau(c),c}-A^{\mathrm{oracle}}_c,\\
 D_c&=A^{\mathrm{adapt}}_{T,c}-A^{\mathrm{adapt}}_{\tau(c),c},\\
 Q_c&=A^{\min}_{T,c}-A^{\mathrm{oracle}}_c.
\end{aligned}
\end{equation}
The final performance gap between Adapted DNE and Separate Min then equals
\begin{equation}
 A^{\mathrm{adapt}}_{T,c}-A^{\min}_{T,c}=G_c+D_c-Q_c.
 \label{eq:decomposition}
\end{equation}
These terms compare adaptation with preservation at acquisition and at the end of the stream. By contrast, the Forgetting Measure (FM) compares a detector with its own earlier best score, so it does not reveal its initial detection quality. For older categories $\mathcal O=\{c:\tau(c)<T\}$, which have at least one later task,
\begin{equation}
 \mathrm{FM}=\frac{1}{|\mathcal O|}\sum_{c\in\mathcal O}
 \left[\max_{\tau(c)\le s<T}A_{s,c}-A_{T,c}\right],
 \label{eq:fm}
\end{equation}
where negative terms are retained when final performance exceeds the earlier best.

\begin{table*}[t]
\caption{\textbf{Comparisons with preserved detectors.} Each row asks what changes when a specified part of the system is varied. Controls apply within the listed detector family; different families retain their own backbones and evaluators.}
\label{tab:controls}
\centering\footnotesize
\setlength{\tabcolsep}{4pt}
\begin{tabularx}{\textwidth}{>{\raggedright\arraybackslash}p{2.75cm}>{\raggedright\arraybackslash}p{3.75cm}>{\raggedright\arraybackslash}p{4.35cm}>{\raggedright\arraybackslash}X}
\toprule
Research Question & Factor Varied & Held Fixed (Control) & Measured Outcome\\
\midrule
How does routing affect detection? & Selector: UCAD, CLS, GCR, vs.\ true-category oracle & Frozen encoder, reference banks, and patch-scoring rule & Routed vs.\ oracle detection performance\\
What does prompt learning add? & Learned category prompts and prompted reference banks & Oracle category selection and reference count ($K=196$) & Within-category image detection and pixel localization\\
Do initial gains persist across tasks? & Adapted vs.\ frozen encoder; merged vs.\ separate Gaussians & Task sequence and evaluator & Initial acquisition ($G_c$), later change ($D_c$), and final detection\\
How does bank size affect detection? & Stored patch references per category ($K_t$) & Frozen encoder, scoring rule, and learned prompts & Routing accuracy, pixel AP (\Pap{}), and compression FM\\
Can one shared model replace separate ones? & Per-category real, joint real, vs.\ sequential generated replay & ReplayCAD architecture, 200 epochs on each input set, and evaluator & Final detection performance and number of models\\
\bottomrule
\end{tabularx}
\end{table*}

\section{Experiments}
\paragraph{Experimental setup.}
We compare preserved patch banks, detectors trained once per category, and sequentially updated shared models on 15 MVTec AD~\cite{mvtec} and 12 VisA~\cite{visa} categories. Here $T\times n$ denotes $T$ tasks of $n$ categories each. DNE uses three category orders for MVTec $5\times3$ and VisA $12\times1$, with a separate MVTec $15\times1$ stream. IUF and CDAD follow their official category orders, one category per task (\suppref{app:code}).

The UCAD routing and prompt comparisons use 196-reference banks with ImageNet-21k ViT-B/16~\cite{vit} or DINOv3 ViT-B~\cite{dinov3}, while preserved Dinomaly detectors use DINOv2~\cite{dinov2}. Training uses seed 0 except for the full-stream IUF runs, which retain seed 133. UCAD prompts are evaluated at the final 25-epoch checkpoint. Each comparison fixes the components listed in Table~\ref{tab:controls}; detector families retain their native backbones, scoring rules, and evaluators.

We report category-mean image AUROC (\Iauc{}), pixel average precision (\Pap{}), pixel AUROC (\Pauc{}), and AUPRO in percent (\%), with differences and FM in percentage points (pp). IUF \Pap{} uses saved $224\times224$ maps and masks. Bootstrap intervals resample categories and, for DNE, task orders (\suppref{app:protocol}).

\subsection{Preserving and Routing Category Detectors}
\label{sec:experts}
We first ask how far inspection can extend without changing an earlier detector. For each new category, we store normal DINOv3 patch features while keeping the encoder fixed. After all categories have arrived, GCR selects the true bank for every MVTec and VisA test image. The resulting \Iauc{} / \Pap{} is 96.22\% / 57.07\% on MVTec and 82.58\% / 29.94\% on VisA, with earlier predictions unchanged.

\begin{table}[t]
\caption{\textbf{Preserving per-category Dinomaly detectors with GCR routing.} Each detector is trained once upon arrival and kept fixed thereafter. GCR routes all 1,725 MVTec and 2,162 VisA test images to their true detector. Metrics (\%) use Dinomaly's evaluator.}
\label{tab:experts}
\centering\small
\setlength{\tabcolsep}{3.1pt}
\begin{tabular}{lrrrrr}
\toprule
Dataset & Route & \Iauc{} & \Pap{} & \Pauc{} & AUPRO\\
\midrule
MVTec & 100.00 & 99.75 & 68.95 & 98.35 & 94.71\\
VisA & 100.00 & 98.93 & 50.70 & 98.97 & 94.77\\
\bottomrule
\end{tabular}
\end{table}

The same approach also works with trained neural detectors. We train one Dinomaly model per category and keep it fixed during later tasks. GCR routes all 1,725 MVTec and 2,162 VisA test images to their true detector, reaching 99.75\% and 98.93\% \Iauc{}, respectively (Table~\ref{tab:experts}). Thus, both stored patch references and separately trained detectors cover the full category stream.

\paragraph{Detection lost through misrouting.}
\label{sec:selectionresults}
Preserved detectors still depend on choosing an appropriate bank or model. To measure how that choice affects detection, we fix the encoder, 196-reference banks, and scoring rule and change only the selector. With prompt-free ImageNet features, replacing UCAD's bank-to-image rule with GCR recovers 9.03 pp \Pap{} on MVTec and 12.55 pp on VisA, matching true-category oracle routing (Table~\ref{tab:selection}).

\begin{table}[t]
\caption{\textbf{Effect of bank selection on defect localization.} Pixel AP (\Pap{}, \%) and gap to true-category oracle routing ($\Delta$, pp, before rounding) under UCAD evaluation. Prompted ImageNet uses the public attention prefix (\suppref{app:prompt}). Cat.\ acc.\ is routing accuracy (\%). DINOv3--VisA no-prompt gap is $-4.9\times10^{-6}$ pp.}
\label{tab:selection}
\centering\footnotesize
\setlength{\tabcolsep}{3pt}
\begin{tabular}{lrrr}
\toprule
Selector & Cat. acc. & \shortstack{No prompt\\P-AP ($\Delta$)} & \shortstack{With prompt\\P-AP ($\Delta$)}\\
\midrule
\multicolumn{4}{l}{\emph{ImageNet / MVTec}}\\
UCAD & 88.64 & 34.45 (-9.03) & 34.59 (-8.78)\\
GCR, CLS / oracle & 100.00 & 43.47 (0.00) & 43.37 (0.00)\\
\multicolumn{4}{l}{\emph{ImageNet / VisA}}\\
UCAD & 79.46 & 11.06 (-12.55) & 7.41 (-11.48)\\
CLS & 99.81 & 22.05 (-1.55) & 16.33 (-2.56)\\
GCR / oracle & 100.00 & 23.61 (0.00) & 18.89 (0.00)\\
\multicolumn{4}{l}{\emph{DINOv3 / MVTec}}\\
All / oracle & 100.00 & 57.07 (0.00) & 59.05 (0.00)\\
\multicolumn{4}{l}{\emph{DINOv3 / VisA}}\\
UCAD & 99.91 & 29.94 ($\approx0$) & 28.86 (-0.001)\\
GCR, CLS / oracle & 100.00 & 29.94 (0.00) & 28.86 (0.00)\\
\bottomrule
\end{tabular}
\end{table}

\begin{figure}[t]
\centering\includegraphics[width=\columnwidth]{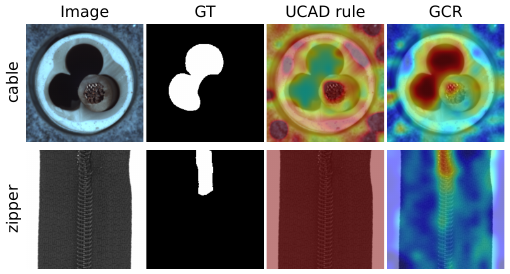}
\caption{\textbf{Selecting the matching reference bank improves localization.} With encoder, 196-reference banks, and prompt-free ImageNet patch scoring fixed, UCAD routes \emph{cable} to \emph{metal-nut} and \emph{zipper} to \emph{toothbrush}. True-bank routing raises per-image \Pap{} from 7.7\% to 94.2\% and 9.8\% to 96.1\%. Examples have the largest AP gaps under the selection rule in \suppref{app:selectionmaps}; each pair shares the oracle map's display scale.}
\label{fig:selectionmaps}
\end{figure}

Figure~\ref{fig:selectionmaps} shows how this happens: metal-nut references assign high scores to normal cable structure, whereas the cable bank localizes the defect. Learning category prompts does not remove the selection error because UCAD first routes with unprompted features and then applies the selected prompt. On ImageNet--MVTec and ImageNet--VisA, it remains 8.78 pp and 11.48 pp below its prompted oracle.

The number of routing errors alone does not determine detection loss. On ImageNet--VisA, CLS misroutes only 4 of 2,162 images (99.81\% accuracy), yet mean \Pap{} falls by 1.55 pp, or 2.56 pp with prompts. Conversely, Separate Min selects the true preserved DNE Gaussian for 90.84\% of MVTec images but matches the oracle within 0.01 pp \Iauc{} (85.03\% vs.\ 85.04\%; \suppref{app:gaussian}). The effect therefore depends on which images are misrouted and how the selected detector scores them.

Feature choice also affects how often routing fails. With DINOv3, all selectors match the oracle on MVTec, and UCAD routes 2,160 of 2,162 VisA images correctly. We further test routing in \emph{Mixed30}, which combines MVTec, VisA, and three medical categories. GCR selects the true category for all 10,900 images, compared with 10,894 for CLS (\suppref{app:routing}).

\subsection{What Continual Learning Adds}
\label{sec:trainingbenefit}
Once a suitable detector can be selected, further learning may still help in three ways: improving detection within a category, making scores more comparable across products, or replacing separate detectors with one shared model. We examine each benefit against its preserved counterpart.

\paragraph{Improving detection within a known category.}
To separate prompt learning from routing, we supply the true category and compare frozen prompt-free features with UCAD's learned prompts. The encoder stays fixed, and the same category prompt is applied to test and bank features (Table~\ref{tab:prompts}). On DINOv3--VisA under UCAD's evaluator, prompts raise \Iauc{} from 82.58\% to 87.55\%, but lower \Pap{} from 29.94\% to 28.86\% and \Pauc{} from 83.79\% to 75.93\%.

We then evaluate the same predictions with a common evaluator that changes map resolution, smoothing, and image-score aggregation (\suppref{app:protocol}). Under this evaluator, \Pap{} instead rises from 38.86\% to 41.21\% (\suppref{app:prompt}). Thus, the localization comparison changes with post-processing, even though the underlying predictions are the same.

For the ImageNet comparison, we correct a batch-dependent attention-prefix operation and retrain the prompts; Table~\ref{tab:selection} retains the public operation for the selector comparison (\suppref{app:prompt}). Both \Iauc{} and \Pap{} fall after prompt training on ImageNet--VisA. Taken together, these results show that prompt learning can affect image detection and pixel localization differently, with gains that depend on the features and evaluator.

\begin{table}[t]
\caption{\textbf{Frozen features vs.\ learned prompts under oracle routing.} Entries report \Iauc{} / \Pap{} (\%) with 196-patch banks and UCAD evaluation. ImageNet uses the corrected attention prefix and retraining (\suppref{app:prompt} also reports public code).}
\label{tab:prompts}
\centering\small
\setlength{\tabcolsep}{3pt}
\begin{tabular}{lrr}
\toprule
Encoder / dataset & Frozen I / P & Prompt I / P\\
\midrule
DINOv3 / MVTec & 96.22 / 57.07 & 97.20 / 59.05\\
DINOv3 / VisA & 82.58 / 29.94 & 87.55 / 28.86\\
ImageNet / MVTec & 90.26 / 43.47 & 90.16 / 42.12\\
ImageNet / VisA & 78.48 / 23.61 & 72.49 / 20.54\\
\bottomrule
\end{tabular}
\end{table}

\begin{figure*}[t]
\centering\includegraphics[width=\textwidth]{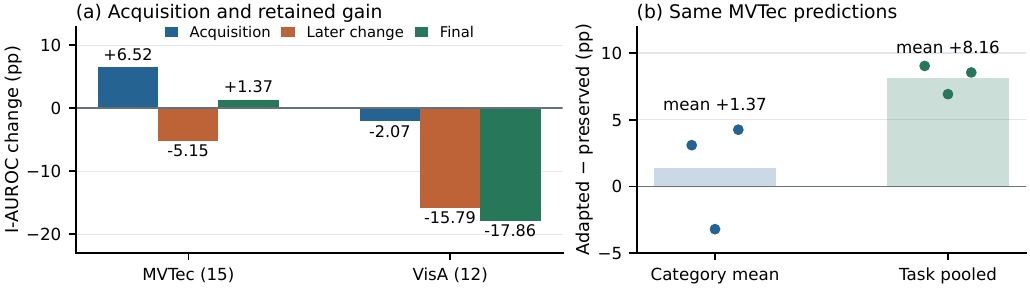}
\caption{\textbf{Tracking initial gains and subsequent changes in DNE.} (a) Decomposition of the final gap between Adapted DNE and the preserved minimum-distance baseline (\emph{Separate Min}) into acquisition gain ($G_c$), later change ($D_c$), and routing ($Q_c$) (Eq.~\eqref{eq:decomposition}). (b) Final MVTec gaps on the same predictions under category-mean vs.\ task-pooled AUROC across three orders (scores in Table~\ref{tab:retained}).}
\label{fig:dne}
\end{figure*}

\paragraph{Comparing anomaly scores across categories.}
Separate detectors can assign scores on different scales. Category-mean AUROC compares normal and defective images within each product, so it does not measure whether scores can be compared across products. We therefore pool images across the three categories within each MVTec task ($5\times3$) and evaluate the same final DNE predictions. Adapted DNE reaches 80.12\% task-pooled AUROC, compared with 71.96\% for Separate Min, and leads on all three orders (Figure~\ref{fig:dne}b). This 8.16 pp gap exceeds the 1.37 pp category-mean gap. Pooling adds cross-product comparison pairs and changes their weighting; under this evaluation, adaptation provides a larger advantage.

\paragraph{Replacing separate detectors with one shared model.}
\label{sec:replayresults}
A shared detector could reduce the state stored as categories grow. We test this with three ReplayCAD configurations using the same architecture and evaluator (Table~\ref{tab:replay}). \emph{Independent real} trains one model per category on real images and uses oracle routing. \emph{Joint real} trains one model on all real images together, while \emph{Generated replay} updates one model sequentially with current real and synthetic historical images. Each configuration trains for 200 epochs on its respective inputs; the input sets and numbers of updates differ.

\begin{table}[t]
\caption{\textbf{Per-category models, joint real training, and generated replay in ReplayCAD} (200 epochs, official evaluator). \emph{Independent real} uses oracle routing; \emph{Joint real} accesses all real normal images jointly; \emph{Generated replay} updates one model sequentially with synthetic historical inputs.}
\label{tab:replay}
\centering\small
\setlength{\tabcolsep}{3.5pt}
\begin{tabular}{lrrrr}
\toprule
Configuration & Models & \Iauc{} & \Pap{} & \Pauc{}\\
\midrule
\multicolumn{5}{l}{\emph{MVTec}}\\
Independent real & 15 & 95.65 & 56.19 & 98.01\\
Joint real & 1 & 98.89 & 56.94 & 98.06\\
Generated replay & 1 & 94.27 & 51.51 & 96.62\\
\multicolumn{5}{l}{\emph{VisA}}\\
Joint real & 1 & 95.94 & 44.25 & 99.03\\
Generated replay & 1 & 88.28 & 41.96 & 98.05\\
\bottomrule
\end{tabular}
\end{table}

On MVTec, Joint real replaces 15 models with one while raising \Iauc{} from 95.65\% to 98.89\% and \Pap{} from 56.19\% to 56.94\%. It obtains this benefit with joint access to real historical images. Generated replay instead finishes at 94.27\% \Iauc{} and 51.51\% \Pap{}, below both the independent and joint real models. Its gaps to Joint real are $-4.62$ pp [$-7.12$, $-2.40$] and $-5.43$ pp [$-9.08$, $-2.03$], respectively. VisA shows the same direction, with 88.28\% \Iauc{} for generated replay versus 95.94\% for joint real training. Generated replay therefore does not recover the shared-model performance obtained with real historical inputs in these comparisons.

\subsection{Do Initial Learning Gains Persist Across Tasks?}
\label{sec:dneresults}
An improvement at acquisition is useful only to the extent that it remains after later tasks. We therefore track the same DNE categories from their first learning step to the final task and compare them with preserved detectors across three orders (Table~\ref{tab:retained}).

\begin{table}[t]
\caption{\textbf{Initial and final DNE detection across streams} (mean \Iauc{}, \%, over three orders). \emph{Oracle} and \emph{Min} use frozen features with preserved Gaussians; \emph{Acquired} and \emph{Final} report Adapted DNE at acquisition and after task $T$. $\Delta$ is Adapted Final minus Preserved Min (pp, before rounding; full decomposition in \suppref{app:decomposition}).}
\label{tab:retained}
\centering\small
\setlength{\tabcolsep}{3pt}
\begin{tabular}{lrrrrr}
\toprule
& \multicolumn{2}{c}{Preserved} & \multicolumn{2}{c}{Adapted DNE} & \\
\cmidrule(lr){2-3}\cmidrule(lr){4-5}
Categories & Oracle & Min & Acquired & Final & $\Delta$\\
\midrule
MVTec: older 12 & 86.42 & 86.41 & 92.56 & 86.12 & $-0.30$\\
MVTec: all 15 & 85.04 & 85.03 & 91.56 & 86.41 & $+1.37$\\
VisA: all 12 & 79.38 & 79.38 & 77.31 & 61.52 & $-17.86$\\
\bottomrule
\end{tabular}
\end{table}

\begin{table*}[t]
\caption{\textbf{Initial (Acquired) and Final scores alongside the Forgetting Measure (FM).} Single-category streams use official orders, with seed 0 for CDAD and 133 for IUF. Acquired and Final average all categories; FM (pp) compares older categories with their pre-final peak. IUF \Pap{} uses saved maps; CDAD paper-basis variants appear in \suppref{app:code}.}
\label{tab:fullstreams}
\centering\small
\setlength{\tabcolsep}{5pt}
\begin{tabular}{llrrrrrrr}
\toprule
& & \multicolumn{3}{c}{I-AUROC} & \multicolumn{3}{c}{P-AP} & P-AUROC\\
\cmidrule(lr){3-5}\cmidrule(lr){6-8}
Dataset & Implementation & Acquired & Final & FM & Acquired & Final & FM & Final\\
\midrule
MVTec $15\times1$ & CDAD / released code & 91.66 & 63.34 & 30.34 & 52.97 & 16.39 & 39.20 & 74.55\\
& IUF / released code & 75.72 & 73.99 & 3.87 & 23.82 & 17.26 & 7.08 & 80.16\\
\midrule
VisA $12\times1$ & CDAD / released code & 84.93 & 55.65 & 32.56 & 31.16 & 7.22 & 26.11 & 88.16\\
& IUF / released code & 57.93 & 59.84 & 0.48 & 3.86 & 3.33 & 0.79 & 84.29\\
\bottomrule
\end{tabular}
\end{table*}

On the 12 MVTec categories introduced before the final task, Adapted DNE reaches 92.56\% \Iauc{} at acquisition, 6.14 pp above the preserved oracle's 86.42\%. Later tasks reduce it by 6.44 pp to 86.12\%, leaving it 0.30 pp below Separate Min (86.41\%). Keeping the initial encoder and category Gaussians would therefore give nearly the same final accuracy on these earlier categories.

The last three MVTec categories have no later training interval, so they contribute only acquisition gains to the all-category mean. Including them gives Adapted DNE a 1.37 pp advantage over Separate Min: Eq.~\eqref{eq:decomposition} separates $+6.52$ pp at acquisition, $-5.15$ pp during later tasks, and $+0.01$ pp from oracle-to-Min selection (Figure~\ref{fig:dne}a). The final difference varies across orders ($+3.09$, $-3.21$, and $+4.25$ pp).

On VisA, adaptation starts below the preserved oracle (77.31\% vs.\ 79.38\%) and falls further to 61.52\%, ending 17.86 pp below preservation. Together, these trajectories show why acquisition and final performance answer different questions: an early gain can be offset by later losses, while some streams begin with no gain to retain.

\begin{figure*}[t]
\centering\includegraphics[width=\textwidth]{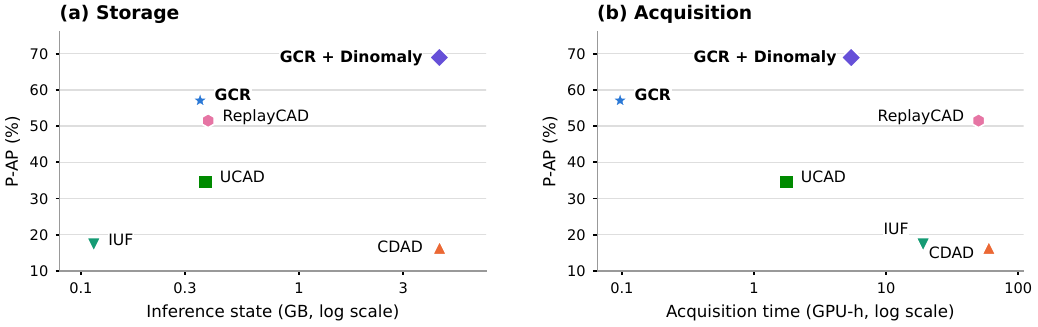}
\caption{\textbf{Detection performance vs.\ storage and acquisition time.} Final MVTec pixel AP (\Pap{}, 15 categories) against (a) stored inference state and (b) stream acquisition time on one RTX 4090 under native evaluators (\suppref{app:cost}; UCAD counts prompt training only, excluding SAM masks and bank registration; DNE appears in the supplementary \Iauc{} plot).}
\label{fig:resources}
\end{figure*}

\paragraph{Interpreting forgetting alongside initial performance.}
\label{sec:fullstreams}
CDAD and IUF show why a forgetting score needs this acquisition context (Table~\ref{tab:fullstreams}). CDAD's MVTec \Iauc{} starts at 91.66\%, but falls on all 14 older MVTec and 11 older VisA categories. Its \Pap{} falls from 52.97\% to 16.39\% on MVTec and from 31.16\% to 7.22\% on VisA. IUF has much smaller VisA FM (0.48 pp \Iauc{}, 0.79 pp \Pap{}), yet its \Pap{} starts at 3.86\% and ends at 3.33\%. Low forgetting here accompanies weak localization from the start.

A small mean can also combine changes in opposite directions. For IUF--VisA, five categories contribute positive FM terms totaling $+3.22$ pp after averaging over the 11 older categories, while six contribute $-2.74$ pp. Their sum leaves 0.48 pp FM, although individual categories change substantially (\suppref{app:qualitative}).

\paragraph{Updating normal statistics with a fixed encoder.}
Weights are not the only stored information that can change. In Frozen--Merged DNE, we fix the initial encoder and retain its feature augmentation, but re-estimate a shared Gaussian from current features and samples from past Gaussians (\suppref{app:gaussian}). Separate Oracle retains each Gaussian and has 0.00 pp FM. Frozen--Merged instead gives 0.77 pp [0.10, 1.47] on MVTec and 1.80 pp [0.94, 2.63] on VisA, with positive FM in all 15 order-by-sampling runs per dataset. Thus, updating the stored normal distribution changes predictions even when the features stay fixed.

\subsection{Resource Costs and Limits of Preservation}
\label{sec:resources}
Keeping earlier detectors avoids retraining them, but requires more stored state as categories arrive (Figure~\ref{fig:resources}). A frozen DINOv3 encoder with 15 MVTec banks occupies 352 MB, adding 0.602 MB per category. Registering all banks takes 0.097 GPU-h, about 23 s per category on one RTX 4090. By comparison, 15 preserved Dinomaly detectors occupy 4.39 GB, add about 246 MB per category, and require 5.443 GPU-h for training and registration.

\paragraph{Restricting reference-bank storage.}
\label{sec:budgetresults}
To limit bank growth, we cap total references at $B$ and reduce each MVTec bank at step $t$ to $K_t=\min(196,\lfloor B/t\rfloor)$. The encoder, scoring rule, and prompts stay fixed, with separate $B$ budgets for routing and scoring banks in prompted UCAD (\suppref{app:budget}). Reducing each bank from 196 to 13 references at $t=15$ lowers \Pap{} from 57.07\% to 28.45\% ($-28.62$ pp), with 13.69 pp FM despite 100\% true-category routing. Correct selection therefore cannot make up for the removed scoring references.

Conversely, banks built with 1,024 references from the outset reach 98.27\% \Iauc{} and 59.89\% \Pap{} with frozen DINOv3 and 100\% routing (Supplementary Table~\ref{tab:largerbank}). Their 15-bank storage rises from about 9.03 MB to 47.2 MB. These larger banks stay fixed across the stream, whereas the compression experiment changes an already stored detector.

\paragraph{Inference latency and shared-model costs.}
We time inference from file reading through routing and scoring to CPU map output on one RTX 4090 (FP32, batch 1, 233 images $\times 3$ runs after warm-up, all 15 detectors resident). Median / p95 latency is 81.6 / 109.4 ms for patch banks and 150.6 / 161.0 ms for preserved Dinomaly. The shared models have different storage and acquisition costs: IUF uses 115 MB and 19.084 GPU-h; ReplayCAD uses 383 MB at inference, excluding its generator, and 50.13 GPU-h to acquire; CDAD uses 4.40 GB and 59.99 GPU-h (\suppref{app:cost}).

\section{Discussion}
\paragraph{When to revisit an already acquired detector.}
A new product adds a category to inspect, but need not change the normal appearance of earlier products. Keeping their detectors fixed therefore provides a direct reference for what repeated training adds. In our comparisons, further learning can improve within-category detection, performance when scores are pooled across products, or the ability to use one shared model. These are different benefits, and each depends on the features, evaluation, and data available for training.

\paragraph{Distinguishing initial gains from long-term retention.}
Acquisition performance shows what the detector learns initially, while later trajectories show how much remains. DNE gains accuracy on the older MVTec categories at acquisition, but later losses offset that gain. IUF illustrates the other case: low forgetting can accompany weak initial detection.

Fixed weights alone also do not guarantee fixed predictions. The chosen detector and its normal references still determine the score: routing to another category, re-estimating a shared Gaussian, or removing patch references can change predictions without updating the encoder. Preservation therefore concerns the stored scoring information as well as the learned parameters.

\paragraph{Resource trade-offs and scope of preservation.}
Separate neural detectors grow with the number of categories, while patch banks add relatively little state beside the shared encoder. Shrinking all 15 MVTec banks from 196 to 13 references saves only 8.43 MB beside a 343 MB encoder, yet sharply reduces localization. Reaching a 115 MB total budget would require a smaller or compressed backbone as well as compact banks.

Joint real-image training offers another way to reduce model count: on MVTec, it replaces separate detectors with one stronger model. Generated replay instead uses synthetic historical inputs and reaches lower accuracy. Our streams keep earlier product appearances stationary and provide category labels during training. Changes within an existing product would create a further reason to update its detector.

\section{Conclusion}
On MVTec AD and VisA, building each category's detector once and routing test images with GCR extends inspection while preserving earlier predictions. Further learning helps in some comparisons, but initial gains can disappear during later tasks, and low forgetting can coexist with weak detection. The value of a continual update is therefore the detection or resource benefit it retains beyond adding the new category.

\label{lastmainpage}
\section*{LLM Usage Statement}
OpenAI ChatGPT and Codex were used to assist with literature searches, writing and debugging aggregation and visualization scripts, and prose editing of the manuscript. The authors take full responsibility for the study design, experimental execution, data analysis, and the final manuscript.

{\small
\bibliographystyle{ieeenat_fullname}
\bibliography{references}
}
\label{lastsubmissionpage}
\clearpage
\appendix
\setcounter{figure}{0}
\setcounter{table}{0}
\setcounter{equation}{0}
\renewcommand{\thefigure}{S\arabic{figure}}
\renewcommand{\thetable}{S\arabic{table}}
\renewcommand{\theequation}{S\arabic{equation}}
\renewcommand{\theHfigure}{supp.\arabic{figure}}
\renewcommand{\theHtable}{supp.\arabic{table}}
\renewcommand{\theHequation}{supp.\arabic{equation}}
\section{What Is Held Fixed in Each Comparison?}
\label{app:protocol}
Preserved detectors provide a reference for what later learning adds. The comparisons below change the routing rule, normal representation, evaluator, or memory budget and state which components remain fixed. We first describe the shared evaluation conditions, then examine the changes in routing, detection, and resource use.

\paragraph{Candidate pool and preserved-detector evaluation.}
Each task supplies labeled normal images for newly introduced categories. At inference, the router chooses among all categories seen so far, breaking ties by a fixed category order. Test category labels are used to compute metrics and to select the true category under oracle routing. Preserved Dinomaly detectors are evaluated after all 15 MVTec or 12 VisA categories have been registered. For uncompressed UCAD banks and preserved prompts, oracle and GCR FM are both zero across all eight evaluated settings.

\paragraph{Converting detector outputs into anomaly scores and maps.}
Routing comparisons use the same anomaly scoring and map processing for every selector. We also compare two evaluators to measure how processing changes the reported scores. UCAD takes the maximum patch distance before upsampling as the image score, resizes maps to $224\times224$ with bilinear interpolation, and applies Gaussian filtering with $\sigma=4$. Ground-truth masks use the public evaluator's resize, center crop, and integer conversion. The common evaluator instead uses $256\times256$ maps, a $5\times5$ Gaussian kernel with $\sigma=4$, and the mean of the top 655 pixels as the image score.

Dinomaly uses its native evaluator, with batch size 16, $256\times256$ maps, and $\sigma=4$ smoothing. ReplayCAD also uses its official detector evaluator. Each learning or routing comparison keeps these operations fixed; the evaluator comparison applies both sets of operations to the same predictions.

\paragraph{Metrics, aggregation, and bootstrap intervals.}
Category-mean metrics are computed separately for each product and averaged equally across categories. Task-pooled AUROC instead combines images from all categories within a task before computing AUROC, adding comparisons between products. Section~\ref{app:prompt} applies both measures to the same predictions.

FM compares each older category's final score with its best pre-final score and retains negative terms when the final score is higher. Backward transfer (BWT) averages final-minus-acquisition differences over the same older categories, while acquisition and final means include all categories. Category trajectories therefore distinguish changes in the detector from changes in the set being averaged. UCAD and DNE uncertainty intervals use 4,000 paired bootstrap resamples over categories and, for DNE, task orders.

\paragraph{Baseline checkpoints and training conditions.}
Table~\ref{tab:published} places the published UCAD and ReplayCAD results beside the executions used here. UCAD uses the final 25-epoch checkpoint. Section~\ref{app:prompt} compares the public ImageNet attention-prefix operation with a corrected version after retraining. ReplayCAD joint real-image training and sequential generated replay are separate conditions because their training inputs and schedules differ.
\begin{table}[!htbp]
\caption{\textbf{Published results and matched comparison runs.} Entries report \Iauc{} / \Pap{} (\%). UCAD uses the public ImageNet prefix and selector at the final 25-epoch checkpoint without test-AUROC early stopping. ReplayCAD uses the official rerun. Controlled component comparisons use matched predictions from these runs.}
\label{tab:published}
\centering\small
\setlength{\tabcolsep}{3pt}
\begin{tabular}{lrr}
\toprule
Method / data & Published I / P & This execution I / P\\
\midrule
UCAD / MVTec & 93.0 / 45.6 & 88.76 / 34.59\\
ReplayCAD / MVTec & 94.8 / 53.7 & 94.27 / 51.51\\
ReplayCAD / VisA & 90.3 / 41.5 & 88.28 / 41.96\\
\bottomrule
\end{tabular}

\end{table}

\section{Can the Appropriate Detector Be Selected as Candidates Grow?}
\label{app:routing}
A stored detector can inspect an unlabeled image only after the system selects it. We therefore compare routing rules with the detectors fixed, increase the number of candidate categories, and inspect the anomaly maps produced by incorrect bank selection.

\subsection{Comparing selectors with fixed detectors}
Table~\ref{tab:twoeval} compares UCAD's bank-to-image rule, CLS, and GCR with oracle selection under both evaluators from Section~\ref{app:protocol}. Within each row group, all selectors use the same encoder, banks, prompts, and scoring rule. Table~\ref{tab:selectionfull} gives routing accuracies and paired 95\% bootstrap intervals for the performance gaps to oracle selection.
\begin{table*}[!htbp]
\caption{\textbf{Comparing routing methods under both evaluators.} Entries report \Iauc{} / \Pap{} (\%), with $\Delta$ denoting routed minus oracle performance (pp, computed before rounding). Within each condition, the encoder, reference banks, prompts, and scoring rule are held fixed. GCR matches oracle routing across all conditions. ImageNet--VisA Common evaluation uses aligned masks across all selectors and prompt settings (prompt-on GCR vs.\ UCAD \Pap{} is 19.52\% vs.\ 8.48\%). Prompt-on ImageNet uses the public prefix; Table~\ref{tab:promptfix} reports the corrected retraining.}
\label{tab:twoeval}
\centering\footnotesize
\setlength{\tabcolsep}{3pt}
\begin{tabular}{lllrrrr}
\toprule
Encoder / data & Prompt & Selector & Common I / P & Common $\Delta$I / $\Delta$P & UCAD I / P & UCAD $\Delta$I / $\Delta$P\\
\midrule
ImageNet / MVTec & off & UCAD & 88.76 / 41.44 & -2.31 / -10.57 & 88.20 / 34.45 & -2.06 / -9.03\\
ImageNet / MVTec & off & CLS & 91.07 / 52.02 & 0.00 / 0.00 & 90.26 / 43.47 & 0.00 / 0.00\\
ImageNet / MVTec & off & GCR / oracle & 91.07 / 52.02 & 0.00 / 0.00 & 90.26 / 43.47 & 0.00 / 0.00\\
ImageNet / MVTec & on & UCAD & 89.69 / 41.43 & -1.91 / -10.48 & 88.76 / 34.59 & -1.69 / -8.78\\
ImageNet / MVTec & on & CLS & 91.59 / 51.92 & 0.00 / 0.00 & 90.45 / 43.37 & 0.00 / 0.00\\
ImageNet / MVTec & on & GCR / oracle & 91.59 / 51.92 & 0.00 / 0.00 & 90.45 / 43.37 & 0.00 / 0.00\\
ImageNet / VisA & off & UCAD & 75.67 / 13.08 & -5.13 / -13.19 & 74.20 / 11.06 & -4.28 / -12.55\\
ImageNet / VisA & off & CLS & 80.79 / 25.10 & 0.00 / -1.16 & 78.48 / 22.05 & +0.00 / -1.55\\
ImageNet / VisA & off & GCR / oracle & 80.79 / 26.26 & 0.00 / 0.00 & 78.48 / 23.61 & 0.00 / 0.00\\
ImageNet / VisA & on & UCAD & 69.66 / 8.48 & -3.84 / -11.04 & 68.77 / 7.41 & -3.81 / -11.48\\
ImageNet / VisA & on & CLS & 73.49 / 17.74 & 0.00 / -1.78 & 72.58 / 16.33 & +0.00 / -2.56\\
ImageNet / VisA & on & GCR / oracle & 73.49 / 19.52 & 0.00 / 0.00 & 72.58 / 18.89 & 0.00 / 0.00\\
DINOv3 / MVTec & off & UCAD & 95.95 / 59.90 & 0.00 / 0.00 & 96.22 / 57.07 & 0.00 / 0.00\\
DINOv3 / MVTec & off & CLS & 95.95 / 59.90 & 0.00 / 0.00 & 96.22 / 57.07 & 0.00 / 0.00\\
DINOv3 / MVTec & off & GCR / oracle & 95.95 / 59.90 & 0.00 / 0.00 & 96.22 / 57.07 & 0.00 / 0.00\\
DINOv3 / MVTec & on & UCAD & 97.16 / 62.20 & 0.00 / 0.00 & 97.20 / 59.05 & 0.00 / 0.00\\
DINOv3 / MVTec & on & CLS & 97.16 / 62.20 & 0.00 / 0.00 & 97.20 / 59.05 & 0.00 / 0.00\\
DINOv3 / MVTec & on & GCR / oracle & 97.16 / 62.20 & 0.00 / 0.00 & 97.20 / 59.05 & 0.00 / 0.00\\
DINOv3 / VisA & off & UCAD & 82.45 / 38.83 & -0.11 / -0.03 & 82.49 / 29.94 & -0.09 / $\approx0$\\
DINOv3 / VisA & off & CLS & 82.56 / 38.86 & 0.00 / 0.00 & 82.58 / 29.94 & 0.00 / 0.00\\
DINOv3 / VisA & off & GCR / oracle & 82.56 / 38.86 & 0.00 / 0.00 & 82.58 / 29.94 & 0.00 / 0.00\\
DINOv3 / VisA & on & UCAD & 85.64 / 41.08 & -0.04 / -0.13 & 87.48 / 28.86 & -0.07 / -0.001\\
DINOv3 / VisA & on & CLS & 85.67 / 41.21 & 0.00 / 0.00 & 87.55 / 28.86 & 0.00 / 0.00\\
DINOv3 / VisA & on & GCR / oracle & 85.67 / 41.21 & 0.00 / 0.00 & 87.55 / 28.86 & 0.00 / 0.00\\
\bottomrule
\end{tabular}

\end{table*}
\begin{table*}[!htbp]
\caption{\textbf{Detection loss from routing errors with fixed detectors.} Routing accuracy (Route), oracle scores, and routed scores are in \%; $\Delta$ is routed minus oracle (pp) with paired 95\% bootstrap intervals. All banks use 196 references under UCAD evaluation. Conditions where GCR and CLS match the oracle exactly are shown in Table~\ref{tab:twoeval}.}
\label{tab:selectionfull}
\centering\footnotesize
\setlength{\tabcolsep}{3pt}
\begin{tabular}{lllrlrrr}
\toprule
Encoder / data & Prompt & Selector & Route & Metric & Oracle & Selected & $\Delta$ [95\% CI]\\
\midrule
ImageNet / MVTec & off & UCAD & 88.64 & I-AUROC & 90.26 & 88.20 & -2.06 [-5.05, +0.24]\\
ImageNet / MVTec & off & UCAD & 88.64 & P-AP & 43.47 & 34.45 & -9.03 [-14.77, -3.72]\\
ImageNet / MVTec & on & UCAD & 88.64 & I-AUROC & 90.45 & 88.76 & -1.69 [-3.88, +0.17]\\
ImageNet / MVTec & on & UCAD & 88.64 & P-AP & 43.37 & 34.59 & -8.78 [-14.37, -3.57]\\
ImageNet / VisA & off & UCAD & 79.46 & I-AUROC & 78.48 & 74.20 & -4.28 [-8.19, -1.22]\\
ImageNet / VisA & off & UCAD & 79.46 & P-AP & 23.61 & 11.06 & -12.55 [-24.23, -2.83]\\
ImageNet / VisA & off & CLS & 99.81 & I-AUROC & 78.48 & 78.48 & +0.00 [0.00, +0.00]\\
ImageNet / VisA & off & CLS & 99.81 & P-AP & 23.61 & 22.05 & -1.55 [-4.66, 0.00]\\
ImageNet / VisA & on & UCAD & 79.46 & I-AUROC & 72.58 & 68.77 & -3.81 [-8.27, -0.17]\\
ImageNet / VisA & on & UCAD & 79.46 & P-AP & 18.89 & 7.41 & -11.48 [-23.07, -1.64]\\
ImageNet / VisA & on & CLS & 99.81 & I-AUROC & 72.58 & 72.58 & +0.00 [0.00, +0.01]\\
ImageNet / VisA & on & CLS & 99.81 & P-AP & 18.89 & 16.33 & -2.56 [-7.68, 0.00]\\
DINOv3 / MVTec & off & UCAD & 100.00 & I-AUROC & 96.22 & 96.22 & 0.00 [0.00, 0.00]\\
DINOv3 / MVTec & off & UCAD & 100.00 & P-AP & 57.07 & 57.07 & 0.00 [0.00, 0.00]\\
DINOv3 / MVTec & on & UCAD & 100.00 & I-AUROC & 97.20 & 97.20 & 0.00 [0.00, 0.00]\\
DINOv3 / MVTec & on & UCAD & 100.00 & P-AP & 59.05 & 59.05 & 0.00 [0.00, 0.00]\\
DINOv3 / VisA & off & UCAD & 99.91 & I-AUROC & 82.58 & 82.49 & -0.09 [-0.27, 0.00]\\
DINOv3 / VisA & off & UCAD & 99.91 & P-AP & 29.94 & 29.94 & $-4.9\!\times\!10^{-6}$ [$-1.5\!\times\!10^{-5}$, 0]\\
DINOv3 / VisA & on & UCAD & 99.91 & I-AUROC & 87.55 & 87.48 & -0.07 [-0.21, 0.00]\\
DINOv3 / VisA & on & UCAD & 99.91 & P-AP & 28.86 & 28.86 & -0.001 [-0.004, 0.000]\\
\bottomrule
\end{tabular}

\end{table*}
On ImageNet--VisA, CLS misroutes four of 2,162 test images (99.81\% routing accuracy) but loses 1.55 pp \Pap{} under UCAD evaluation. With DINOv3--MVTec, all three selectors route every image correctly and match the oracle. Feature choice thus affects how often routing fails, while the images and reference banks involved determine the resulting detection loss.

\subsection{Scaling routing to larger candidate pools}
\begin{table}[!htbp]
\caption{\textbf{Category routing accuracy as the candidate pool expands.} DINOv3 final-pool routing. \emph{Unified18} combines the 15 MVTec categories with three medical categories (brain, liver, and retinal images); \emph{Mixed30} adds the 12 VisA categories.}
\label{tab:mixedrouting}
\centering\small
\setlength{\tabcolsep}{4pt}
\begin{tabular}{lrr}
\toprule
Candidate pool & GCR correct / total & CLS correct / total\\
\midrule
MVTec (15) & 1,725 / 1,725 & 1,725 / 1,725\\
VisA (12) & 2,162 / 2,162 & 2,162 / 2,162\\
Unified18 & 8,738 / 8,738 & 8,732 / 8,738\\
Mixed30 & 10,900 / 10,900 & 10,894 / 10,900\\
\bottomrule
\end{tabular}

\end{table}
To test routing with more candidates, Mixed30 combines the 15 MVTec categories, the 12 VisA categories, and three medical categories (Table~\ref{tab:mixedrouting}). VisA includes visually similar categories such as \emph{macaroni1/2} and \emph{pcb1--4}. GCR selects the true category for all 10,900 Mixed30 images, whereas CLS routes six brain images to the liver category.

We next measure how close each router comes to selecting a wrong category. Both GCR and CLS, expressed as $1-\text{cosine similarity}$, prefer smaller distances. We define the \emph{routing margin} as the smallest wrong-category distance minus the true-category distance. Positive values indicate correct routing; negative values indicate an error.

\begin{figure*}[!htbp]
\centering\includegraphics[width=\textwidth]{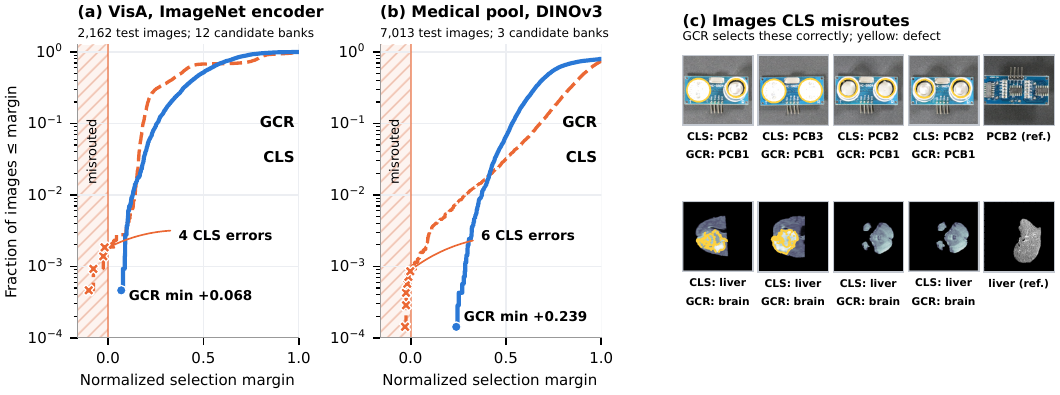}
\caption{\textbf{Routing margin distributions for GCR and CLS.} (a) ImageNet--VisA (12 banks, 196 references each). (b) DINOv3 medical pool (3 banks, 1,024 references each). Curves show the cumulative fraction of test images at or below each normalized margin; negative values correspond to routing errors. GCR maintains positive margins for all images, whereas CLS makes four errors on VisA and six on the medical pool. (c) Qualitative examples of the four VisA and four medical images misrouted by CLS alongside references from the falsely selected categories (yellow contours mark ground-truth defects).}
\label{fig:margins}
\end{figure*}

Because the two selectors have different distance scales, we normalize their margins separately (Figure~\ref{fig:margins}). For each image, we first take the median wrong-minus-true distance over wrong categories, then use the dataset median of those values as the divisor. This common positive divisor within each selector preserves the margin signs and their ordering.
For ImageNet--VisA, the smallest normalized margin is~$+0.068$ with GCR and $-0.102$ with CLS. For the separate 7,013-image DINOv3 medical pool, these margins are~$+0.239$ and $-0.030$, respectively; neither pool contains a tie.

Across four arrival orders of the 15 MVTec categories, the final reference banks are byte-identical and the final detection scores are unchanged. With fixed category scores and a fixed tie rule, an image routed correctly against the full candidate set is also routed correctly against any earlier subset containing its true category. Compression changes the banks themselves, so Section~\ref{app:budget} instead evaluates compressed banks at every stage.

\subsection{Effect of misrouted reference banks on anomaly maps}
\label{app:selectionmaps}
Main-paper Figure~\ref{fig:selectionmaps} shows how bank selection changes anomaly maps with the ImageNet ViT-B/16 encoder, 196-reference banks, and prompt-free scoring fixed. UCAD routes \emph{cable} to \emph{metal-nut} and \emph{zipper} to \emph{toothbrush}; their per-image \Pap{} is then 7.684\% and 9.760\%. GCR selects the true banks and reaches 94.197\% and 96.136\%, respectively.

For each category, we select the image with the largest AP gap among images with 1--25\% defect area that UCAD misroutes and GCR routes correctly. Each pair uses the 1st-to-99.5th percentile range of the oracle map for both displays. Thus, the displayed contrast comes from bank selection under a shared color scale.

\section{Do Initial Learning Gains Persist Across Subsequent Tasks?}
\label{app:qualitative}
\label{app:decomposition}
\label{app:code}
To determine how much of an initial gain remains, we follow each category from acquisition to the final task and compare its scores with the preserved detector. Category trajectories show the aggregate change, while maps of the same defect images show how localization changes.

\subsection{Decomposing initial acquisition gains and subsequent changes}
The preserved DNE baseline fixes the initial encoder and stores one Gaussian per category. Adapted DNE instead updates the encoder and Gaussians across tasks, following the official implementation in freezing the patch embedding and classification head after task 1. Table~\ref{tab:decomp} compares the same categories and orders, separating the initial acquisition difference ($G$), later change ($D$), and label-free selection difference ($Q$).
\begin{table*}[!htbp]
\caption{\textbf{Decomposition of DNE performance into initial acquisition gain and subsequent change.} Entries report \Iauc{} (\%) and differences (pp), averaged across categories and three task orders. \emph{Oracle} is preserved Separate Oracle; \emph{Acquired} and \emph{Final} track Adapted DNE; \emph{Min} is preserved Separate Min. By Eq.~\eqref{eq:decomposition}, $G=\mathrm{Acquired}-\mathrm{Oracle}$, $D=\mathrm{Final}-\mathrm{Acquired}$, $-Q=\mathrm{Oracle}-\mathrm{Min}$, and $\mathrm{Final}\ \Delta=\mathrm{Final}-\mathrm{Min}=G+D-Q$ (computed before rounding).}
\label{tab:decomp}
\centering\small
\setlength{\tabcolsep}{3pt}
\begin{tabular}{lrrrrrrrr}
\toprule
Categories & Oracle & Acquired & Final & Min & G & D & $-$Q & Final $\Delta$\\
\midrule
MVTec: all 15 & 85.04 & 91.56 & 86.41 & 85.03 & +6.52 & -5.15 & +0.01 & +1.37\\
MVTec: older 12 & 86.42 & 92.56 & 86.12 & 86.41 & +6.14 & -6.44 & +0.00 & -0.30\\
VisA: all 12 & 79.38 & 77.31 & 61.52 & 79.38 & -2.07 & -15.79 & 0.00 & -17.86\\
VisA: older 11 & 78.28 & 77.02 & 59.79 & 78.28 & -1.26 & -17.23 & 0.00 & -18.49\\
\bottomrule
\end{tabular}

\end{table*}
Across all categories, the acquisition difference $G$ relative to the preserved oracle is $+6.52$ pp [$+3.16$, $+10.72$] on MVTec and $-2.07$ pp [$-4.28$, $+0.14$] on VisA. Among categories introduced before the final task, later change $D$ is $-6.44$ pp [$-12.94$, $-2.03$] for the 12 older MVTec categories and $-17.23$ pp [$-22.51$, $-12.45$] for the 11 older VisA categories. On MVTec, this decline offsets the older categories' $+6.14$ pp initial gain. On VisA, acquisition is already below preservation and later tasks widen the gap.

The separate MVTec $15\times1$ stream gives 85.36\% final \Iauc{}, 6.28 pp FM, and $-5.06$ pp BWT for Adapted DNE. Section~\ref{app:gaussian} examines another source of change by fixing the encoder and updating only the stored Gaussian distribution.

\subsection{Category-wise trajectories and qualitative anomaly maps}
\begin{figure*}[!htbp]
\centering\includegraphics[width=\textwidth]{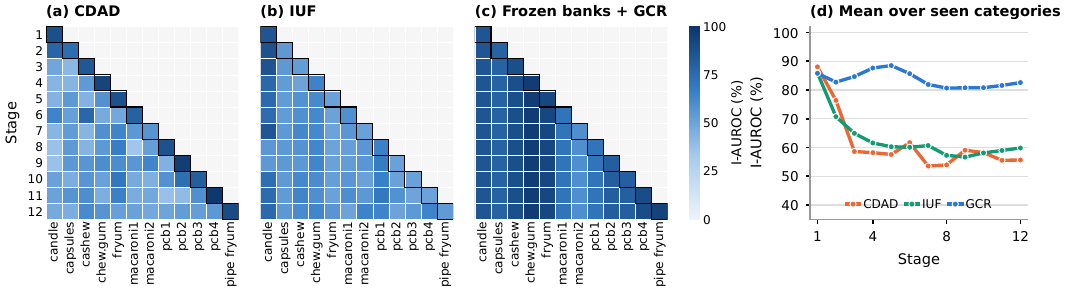}
\caption{\textbf{Category-wise \Iauc{} trajectories on VisA ($12\times1$ stream).} Each column follows a single category from its acquisition step (outlined diagonal) through the final task (bottom row), plotted on a 0--100\% scale matching Figure~\ref{fig:stream}. Right panel shows running stage means over the categories introduced up to each step.}
\label{fig:streamvisa}
\end{figure*}
Category trajectories explain how the means arise (Figure~\ref{fig:streamvisa}). CDAD's \Iauc{} falls from acquisition to the final task on every older category, with median drops of 33.22 pp on MVTec and 34.82 pp on VisA. IUF instead declines on 10 of 14 older MVTec categories and improves on 4; on VisA, 5 decline and 6 improve.

IUF also starts with very different detection quality across categories. Its initial MVTec scores reach 99.68\% on \emph{carpet}, 100.00\% on \emph{leather}, 95.56\% on \emph{tile}, 97.19\% on \emph{wood}, and 80.12\% on \emph{grid}. They are much lower on \emph{cable} (55.98\%), \emph{pill} (52.21\%), \emph{screw} (54.56\%), and \emph{transistor} (52.96\%). A later change of the same size therefore starts from very different detection performance.

On IUF--VisA, positive FM terms contribute $+3.2209$ pp and negative terms contribute $-2.7361$ pp after division by the 11 older categories, leaving 0.4848 pp FM. For example, \emph{candle} drops from 85.90\% to 76.81\%, while \emph{macaroni2} rises from 51.76\% to 65.16\%. Preserved GCR banks instead keep every category column constant, so their running mean changes only when a new category enters the average.
\begin{figure*}[!htbp]
\centering\includegraphics[width=\textwidth]{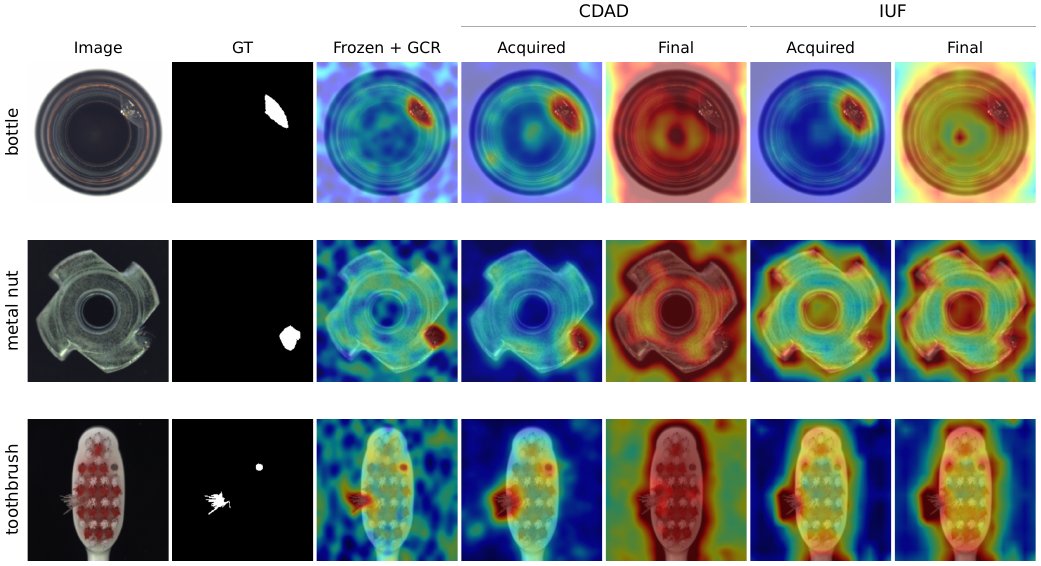}
\caption{\textbf{Anomaly maps at acquisition and after the final task.} MVTec \emph{Bottle}, \emph{Metal Nut}, and \emph{Toothbrush} images compare preserved GCR banks with released-code CDAD and IUF. Examples maximize the CDAD acquisition-to-final AP drop under the selection rule below. Each method/category pair shares the acquisition map's display scale (1st--99.5th percentile).}
\label{fig:qualitativefull}
\label{fig:qualitative}
\end{figure*}
Figure~\ref{fig:qualitativefull} and Table~\ref{tab:qualitativeap} show the same contrast on selected defect images. On \emph{Bottle}, CDAD's per-image \Pap{} falls from 93.27\% to 5.37\%, and IUF's falls from 94.79\% to 70.26\%, while the preserved bank stays at 98.05\%. CDAD also loses localization on \emph{Metal Nut} and \emph{Toothbrush}. IUF starts with only 15.45\% and 24.12\% \Pap{} on these images, so its smaller later change accompanies weak localization from acquisition.
\begin{table}[!htbp]
\caption{\textbf{Per-image pixel AP (\%) for the qualitative examples in Figure~\ref{fig:qualitativefull}.}}
\label{tab:qualitativeap}
\centering\small
\setlength{\tabcolsep}{3pt}
\begin{tabular}{lrrr}
\toprule
Image category & GCR & CDAD acq. / final & IUF acq. / final\\
\midrule
Bottle & 98.05 & 93.27 / 5.37 & 94.79 / 70.26\\
Metal nut & 94.47 & 87.26 / 2.37 & 15.45 / 10.40\\
Toothbrush & 70.92 & 61.70 / 2.78 & 24.12 / 11.30\\
\bottomrule
\end{tabular}

\end{table}
The displayed images have 1--25\% defect area and GCR \Pap{} at or above their category median. Among those images, we select the largest CDAD acquisition-to-final AP drop in each category and render Jet heatmaps at 50\% opacity.

\subsection{Two-stage change after introducing a single new category}
\label{app:twostage}
A two-stage \emph{Bottle} $\rightarrow$ \emph{Cable} stream tests whether a single new category changes an earlier detector (Table~\ref{tab:twostage}). Here $A_{1,1}$ is \emph{Bottle} performance just after learning \emph{Bottle}; $A_{2,1}$ is its performance after learning \emph{Cable}; and $A_{2,2}$ is \emph{Cable} performance at stage 2. Evaluation uses batch size 1 and each method's native scoring rule.
\begin{table}[!htbp]
\caption{\textbf{Change in \emph{Bottle} detection after learning a single new category (\emph{Cable}).} Scores are in \%; $\Delta=A_{2,1}-A_{1,1}$ is in pp, where $A_{1,1}$ is \emph{Bottle} after stage 1, $A_{2,1}$ is \emph{Bottle} after stage 2 (\emph{Cable}), and $A_{2,2}$ is \emph{Cable} at stage 2.}
\label{tab:twostage}
\centering\small
\setlength{\tabcolsep}{2.5pt}
\begin{tabular}{lrrrr}
\toprule
Method / metric & $A_{1,1}$ & $A_{2,1}$ & $A_{2,2}$ & $\Delta$\\
\midrule
IUF / \Iauc{} & 99.52 & 93.89 & 58.34 & $-5.63$\\
IUF / \Pauc{} & 98.20 & 89.15 & 77.54 & $-9.05$\\
ReplayCAD / \Iauc{} & 100.00 & 98.41 & 96.59 & $-1.59$\\
ReplayCAD / \Pap{} & 79.16 & 71.50 & 60.30 & $-7.66$\\
CDAD / \Iauc{} & 98.41 & 89.13 & 76.80 & $-9.29$\\
CDAD / \Pauc{} & 97.94 & 78.47 & 94.66 & $-19.47$\\
CDAD / \Pap{} & 74.05 & 32.77 & 38.16 & $-41.28$\\
CDAD / PRO & 92.74 & 52.17 & 81.72 & $-40.56$\\
\bottomrule
\end{tabular}

\end{table}
Each model uses the same evaluator at both stages. IUF and CDAD retain their public AUROC polarity rule: IUF--\emph{Cable}, for example, scores 58.34\% with reversal and 41.66\% without. For PRO at FPR 0.3, UCAD normalizes by 0.3, while CDAD uses the largest sampled FPR below 0.3. Within these evaluation settings, learning one further category lowers earlier detection in all three methods.

\subsection{Examining the update rules in released implementations}
The trajectories above describe the trained models. To examine the updates that produce them, we also inspect IUF's loss gradients and CDAD's stored projection basis in the public implementations~\cite{iuf,cdad}. Table~\ref{tab:commits} identifies the versions used.
\begin{table}[!htbp]
\caption{\textbf{Public repository versions used in our update analysis.} Commit dates are in UTC; each 40-character SHA is split across two lines.}
\label{tab:commits}
\centering\small
\setlength{\tabcolsep}{3pt}
\begin{tabular}{ll}
\toprule
Repository / date (UTC) & Full commit SHA\\
\midrule
\shortstack[l]{\texttt{jqtangust/IUF}\\2025-02-17} & \href{https://github.com/jqtangust/IUF/commit/f42669d8f60e7de61f4abc02ed0abfb4bcf98a4a}{\shortstack[l]{\texttt{f42669d8f60e7de61f4a}\\\texttt{bc02ed0abfb4bcf98a4a}}}\\
\addlinespace
\shortstack[l]{\texttt{FuNz-0/One-for-More}\\2025-05-07} & \href{https://github.com/FuNz-0/One-for-More/commit/f4eb78841dbfa5612e008570b690072b19a3d9b3}{\shortstack[l]{\texttt{f4eb78841dbfa5612e00}\\\texttt{8570b690072b19a3d9b3}}}\\
\bottomrule
\end{tabular}

\end{table}
\paragraph{IUF decoder gradient flow.}
IUF defines semantic compression loss (SCL) on singular values of decoder features, written as $L_{sc}=\sum_{i=t}^{C}\sigma_i$. The released code detaches three decoder feature tensors before computing SCL. It then normalizes singular values by their sum and averages entries below $0.1/256$ across the three layers. Because of the detach operation, SCL sends no gradient to the decoder; the decoder receives gradients from reconstruction and cross-entropy losses.

\paragraph{CDAD gradient projection basis.}
CDAD stores normal directions as columns of $P$ and updates a gradient matrix $G$ by $G\leftarrow G-GPP^T$. If these columns are orthonormal, $PP^T$ projects onto their span and the update removes gradient components in that subspace. The paper appends orthonormal residual left singular vectors, while the released code appends unnormalized residual activation columns. To see the effect, take an initial basis vector $e_1=(1,0)^T$ and a new activation $a=(1/2,\sqrt{3}/2)^T$. Appending its raw residual gives
\begin{equation}
 P=\begin{pmatrix}1&0\\0&\sqrt{3}/2\end{pmatrix},\qquad
 P^TP=\operatorname{diag}(1,3/4).
\end{equation}
For $G=(0,1)$, this basis leaves $(0,1/4)$ after the update. An orthonormal basis spanning the same 2D space removes the gradient completely. Thus, the column norms as well as the span of $P$ determine how much of the gradient remains.

\paragraph{Comparing released and paper-basis CDAD streams.}
Table~\ref{tab:fullstreamsvariants} compares two CDAD variants. The released version appends residual activations, uses an inter-task energy threshold of 0.5, and trains for 500 epochs per task. The paper-basis version implements Eq.~(6) of~\cite{cdad}: it appends orthonormal left singular vectors with a 0.999 threshold and trains for 500 epochs on task 1 and 100 epochs thereafter. This comparison jointly changes the basis rule, threshold, and training schedule. Both variants decline from acquisition to the final task on MVTec and VisA.

CDAD forms anomaly maps by comparing ResNet-50 features of its VAE and diffusion reconstructions at selected layers. On the first VisA paper-basis checkpoint, the VisA configuration's layers $[0,2,4]$ give 88.06\% / 18.40\% \Iauc{} / \Pap{}, compared with 82.89\% / 13.36\% for layers $[1,2,3]$.

The IUF single-category stream extends the released multi-step configuration to one category per task and uses seed 133. Where no validation split is available, the validation-split feature pass is skipped, retaining the existing feature matrix at that pass. Evaluation uses saved $224\times224$ maps. These execution conditions accompany the released-loss results reported in Table~\ref{tab:fullstreamsvariants}.
\begin{table*}[!htbp]
\caption{\textbf{Released and paper-basis CDAD variants alongside IUF.} Single-category streams use official orders, with seed 0 for CDAD and 133 for IUF. Acquired and Final average all categories; FM (pp) compares older categories with their pre-final peak.}
\label{tab:fullstreamsvariants}
\centering\small
\setlength{\tabcolsep}{5pt}
\begin{tabular}{llrrrrrrr}
\toprule
& & \multicolumn{3}{c}{I-AUROC} & \multicolumn{3}{c}{P-AP} & P-AUROC\\
\cmidrule(lr){3-5}\cmidrule(lr){6-8}
Dataset & Implementation & Acquired & Final & FM & Acquired & Final & FM & Final\\
\midrule
MVTec & CDAD / released code & 91.66 & 63.34 & 30.34 & 52.97 & 16.39 & 39.20 & 74.55\\
$15\times1$ & CDAD / paper Eq.~(6) & 91.63 & 62.74 & 31.88 & 50.92 & 17.96 & 35.32 & 78.48\\
& IUF / released code & 75.72 & 73.99 & 3.87 & 23.82 & 17.26 & 7.08 & 80.16\\
\midrule
VisA & CDAD / released code & 84.93 & 55.65 & 32.56 & 31.16 & 7.22 & 26.11 & 88.16\\
$12\times1$ & CDAD / paper Eq.~(6) & 86.01 & 58.02 & 31.30 & 30.63 & 7.53 & 25.29 & 87.78\\
& IUF / released code & 57.93 & 59.84 & 0.48 & 3.86 & 3.33 & 0.79 & 84.29\\
\bottomrule
\end{tabular}

\end{table*}

\section{Why Can Detection Change While the Encoder Stays Fixed?}
\label{app:gaussian}
\label{app:budget}
The preceding trajectories follow models whose weights change across tasks. Here we fix the encoder and change the stored normal information: either we re-estimate a shared Gaussian or reduce patch banks as categories arrive. These comparisons test whether fixed features alone keep detection unchanged.

\subsection{Re-estimating the normal Gaussian distribution}
DNE scores normalized class-token features by Mahalanobis distance. Frozen--Merged keeps the initial encoder $\Theta_0$ fixed and re-estimates a shared Gaussian from current-task features and samples drawn from historical Gaussians. Separate Oracle and Separate Min instead keep each category's Gaussian $(\mu_c,\Sigma_c)$ unchanged. The former uses the true category; the latter selects the smallest distance among the categories seen so far. All three conditions retain DNE's normal-feature augmentation. For normalized features $z(x)=f^{\mathrm{CLS}}_{\Theta_0}(x)/\|f^{\mathrm{CLS}}_{\Theta_0}(x)\|_2$, we use
\begin{equation}
 \begin{aligned}
 d_c(x)&=\operatorname{Mahalanobis}(z(x);\mu_c,\Sigma_c),\\
 a_t^{\min}(x)&=\min_{c\in\mathcal C_t}d_c(x).
 \end{aligned}
\end{equation}
\begin{table}[!htbp]
\caption{\textbf{Effect of Gaussian re-estimation under a frozen encoder.} Initial (acquired) and final category-mean \Iauc{} (\%) and FM (pp), averaged over three task orders. All frozen conditions use the exact same unchanged encoder.}
\label{tab:dne}
\centering\small
\setlength{\tabcolsep}{4pt}
\begin{tabular}{lrr}
\toprule
Condition / quantity & MVTec $5\times3$ & VisA $12\times1$\\
\midrule
Adapted DNE: acquired & 91.56 & 77.31\\
Adapted DNE: final & 86.41 & 61.52\\
Frozen--Merged: final & 83.06 & 75.30\\
Separate Oracle: final & 85.04 & 79.38\\
Separate Min: final & 85.03 & 79.38\\
\midrule
Adapted DNE: FM & 7.00 & 18.60\\
Frozen--Merged: FM & 0.77 & 1.80\\
Separate Oracle / Min: FM & 0.000 / 0.005 & 0.000 / 0.000\\
Min category accuracy & 90.84 & 99.72\\
\bottomrule
\end{tabular}

\end{table}
Separate Min selects the true category for 90.84\% of MVTec test images but reaches 85.03\% \Iauc{}, close to Separate Oracle's 85.04\% (Table~\ref{tab:dne}). Gaussian selection therefore has little effect on image detection in this setting, despite its category errors. Separate Oracle has zero FM. Separate Min has 0.000 pp on VisA and 0.004873 pp on MVTec because a newly added Gaussian can reduce an earlier image's minimum distance even when its original Gaussian stays fixed.

Frozen--Merged instead changes the shared distribution at each task. With seed-0 features fixed, five Gaussian-sampling streams (S0--S4) for each of three orders give positive FM in all 15 runs per dataset (Table~\ref{tab:streams}). Final VisA \Iauc{} ranges from 74.75\% to 75.76\%, below the 79.38\% from separate preserved Gaussians. The stored distribution can therefore change detection even when the encoder does not change.
\begin{table}[!htbp]
\caption{\textbf{Forgetting Measure (FM, pp) for Frozen--Merged DNE across five Gaussian-sampling streams (S0--S4) and three task orders.}}
\label{tab:streams}
\centering\small
\begin{tabular}{llrrrrr}
\toprule
Dataset & Order & S0 & S1 & S2 & S3 & S4\\
\midrule
MVTec & 1 & 0.93 & 0.82 & 0.91 & 1.06 & 1.32\\
& 2 & 0.21 & 0.37 & 0.68 & 0.12 & 0.56\\
& 3 & 1.17 & 0.56 & 0.94 & 1.67 & 1.29\\
VisA & A & 2.59 & 1.78 & 2.77 & 1.97 & 2.33\\
& B & 2.48 & 1.12 & 2.31 & 2.63 & 2.41\\
& C & 1.45 & 1.72 & 1.22 & 1.48 & 2.01\\
\bottomrule
\end{tabular}

\end{table}

\subsection{Subsampling normal patch reference banks under a fixed budget}
With a fixed total reference budget $B$, earlier banks must shrink as categories arrive. At step $t$, we retain the first $K_t=\min(196,\lfloor B/t\rfloor)$ references in each bank, keeping the encoder, scoring rule, and learned prompts fixed. Prompted settings use separate $B$ budgets for routing and scoring banks. GCR routes all images correctly at the final step for every bank size tested, so the final localization loss reflects reference removal rather than misrouting.
\begin{figure}[!htbp]
\centering
\includegraphics[width=\columnwidth]{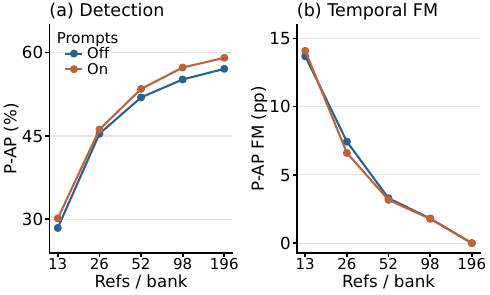}
\caption{\textbf{Effect of stored-order bank compression on DINOv3--MVTec localization.} Final \Pap{} and compression-induced FM as total budget $B$ decreases, while final GCR routing accuracy remains 100\% at every bank size. Learned prompts are held fixed, with separate routing and scoring bank budgets under the prompted setting.}
\label{fig:budget}
\end{figure}
\begin{table*}[!htbp]
\caption{\textbf{Detection performance and compression-induced FM under a fixed total reference budget $B$ (DINOv3--MVTec, stored-order subsampling).} \emph{Off} and \emph{On} denote prompt-free and prompted settings; entries report \Iauc{} / \Pap{} (\%) and FM (pp).}
\label{tab:budgetfull}
\centering\small
\begin{tabular}{rrcccc}
\toprule
$B$ & Final $K$ & Off I / P & On I / P & Off FM I / P & On FM I / P\\
\midrule
2940 & 196 & 96.22 / 57.07 & 97.20 / 59.05 & 0.00 / 0.00 & 0.00 / 0.00\\
1470 & 98 & 94.04 / 55.16 & 96.21 / 57.31 & 2.01 / 1.81 & 0.57 / 1.78\\
784 & 52 & 91.46 / 51.93 & 90.74 / 53.46 & 3.57 / 3.29 & 4.08 / 3.16\\
392 & 26 & 86.86 / 45.39 & 88.05 / 46.15 & 6.07 / 7.44 & 5.30 / 6.61\\
196 & 13 & 80.67 / 28.45 & 76.18 / 30.15 & 7.51 / 13.69 & 14.11 / 14.09\\
\bottomrule
\end{tabular}

\end{table*}
When each bank is reduced from 196 references to 13 at the final task ($B=196$), prompt-free \Pap{} drops from 57.07\% to 28.45\% with 13.69 pp of \Pap{} FM across the stream; with learned prompts, \Pap{} drops from 59.05\% to 30.15\% with 14.09 pp FM (Figure~\ref{fig:budget}, Table~\ref{tab:budgetfull}). Table~\ref{tab:budgetgain} reports the differences between prompted and prompt-free detection at each compression level.
\begin{table*}[!htbp]
\caption{\textbf{Effect of learned prompts across reference-bank sizes.} Entries report \Iauc{} / \Pap{} (\%) and prompted-minus-frozen differences ($\Delta$, pp, computed before rounding) with paired 95\% bootstrap intervals.}
\label{tab:budgetgain}
\centering\small
\setlength{\tabcolsep}{3pt}
\begin{tabular}{rrrrr}
\toprule
Final K & Off I / P & On I / P & $\Delta$I [95\% CI] & $\Delta$P [95\% CI]\\
\midrule
196 & 96.22 / 57.07 & 97.20 / 59.05 & +0.98 [-0.10, +2.35] & +1.98 [-0.72, +5.18]\\
98 & 94.04 / 55.16 & 96.21 / 57.31 & +2.17 [-0.08, +5.19] & +2.15 [-0.47, +5.25]\\
52 & 91.46 / 51.93 & 90.74 / 53.46 & -0.72 [-3.82, +2.03] & +1.53 [-1.05, +4.15]\\
26 & 86.86 / 45.39 & 88.05 / 46.15 & +1.20 [-1.16, +3.65] & +0.76 [-3.04, +4.79]\\
13 & 80.67 / 28.45 & 76.18 / 30.15 & -4.49 [-12.21, +3.17] & +1.70 [-7.35, +11.51]\\
\bottomrule
\end{tabular}

\end{table*}
\begin{table*}[!htbp]
\caption{\textbf{Bank compression via farthest-point sampling within stored banks.} Final GCR and oracle detection scores agree; oracle FM is reported to exclude intermediate-stage routing errors.}
\label{tab:farthest}
\centering\small
\begin{tabular}{rcccc}
\toprule
Final $K$ & Off I / P & On I / P & Oracle FM, off I / P & Oracle FM, on I / P\\
\midrule
196 & 96.22 / 57.07 & 97.20 / 59.05 & 0.00 / 0.00 & 0.00 / 0.00\\
98 & 90.18 / 52.05 & 91.95 / 54.09 & 3.98 / 3.52 & 4.02 / 3.96\\
52 & 88.65 / 48.87 & 90.19 / 49.98 & 4.55 / 4.61 & 3.18 / 4.04\\
26 & 85.26 / 40.37 & 83.92 / 40.65 & 5.62 / 8.44 & 7.21 / 9.63\\
13 & 80.56 / 29.01 & 79.73 / 31.97 & 5.24 / 12.37 & 7.04 / 11.33\\
\bottomrule
\end{tabular}

\end{table*}
Farthest-point sampling within the stored banks gives a similar decline (Table~\ref{tab:farthest}). At $K=13$, final routing is again 100\%, although some images are misrouted at intermediate steps. GCR \Iauc{} / \Pap{} FM is 5.07 / 12.37 pp without prompts and 6.86 / 11.04 pp with prompts. Supplying the true category at every step also gives positive FM, as shown by the oracle columns.

\subsection{Constructing larger reference banks from the outset}
\begin{table}[!htbp]
\caption{\textbf{Detection performance across per-category reference counts $K$ (DINOv3--MVTec).} Routing accuracy is 100\% across all rows. The $K=1{,}024$ condition uses separately registered banks totaling 47.2 MB; prompt learning at $K=1{,}024$ was not evaluated (---).}
\label{tab:largerbank}
\centering\small
\setlength{\tabcolsep}{4pt}
\begin{tabular}{rrr}
\toprule
References $K$ & Frozen I / P & Prompt I / P\\
\midrule
13 & 80.67 / 28.45 & 76.18 / 30.15\\
26 & 86.86 / 45.39 & 88.05 / 46.15\\
52 & 91.46 / 51.93 & 90.74 / 53.46\\
98 & 94.04 / 55.16 & 96.21 / 57.31\\
196 & 96.22 / 57.07 & 97.20 / 59.05\\
\midrule
1,024 & 98.27 / 59.89 & ---\\
\bottomrule
\end{tabular}

\end{table}
We next build banks with 1,024 references per category and keep them at that size throughout the stream (Table~\ref{tab:largerbank}). Frozen DINOv3 with GCR reaches 98.27\% \Iauc{} and 59.89\% \Pap{}, with every image routed correctly. The 15 banks occupy 47.2 MB, compared with about 9.03 MB at $K=196$; including the 342.852 MB encoder gives 390.1 MB in total. The rows at $K\le196$ come from stored-bank compression, while the 1,024-reference banks are registered separately at that size.

\section{When Does Continual Learning Add a Useful Capability?}
\label{app:prompt}
\label{app:replay}
We examine three possible benefits beyond keeping earlier detectors: better detection within a category, better performance when scores are pooled across products, and a single model that replaces separate category detectors.

\subsection{Within-category detection with learned prompts}
Oracle routing supplies the true category, allowing us to compare UCAD's learned prompts with frozen features without selection errors. Table~\ref{tab:prompteval} reports both evaluators. On DINOv3--VisA, prompts raise \Iauc{} under both, but their \Pap{} effect changes from $-1.08$ pp under UCAD evaluation to $+2.35$ pp under the common evaluator. The localization comparison therefore depends on map resolution and smoothing.
\begin{table*}[!htbp]
\caption{\textbf{Comparing frozen and prompted detection under oracle routing across evaluators.} \emph{Off} and \emph{On} report scores without and with learned prompts (\%); $\Delta$ is prompted minus frozen (pp, computed before rounding) with paired 95\% category-bootstrap intervals. ImageNet rows use the public prefix here; corrected retraining is reported in Table~\ref{tab:promptfix}.}
\label{tab:prompteval}
\centering\small
\setlength{\tabcolsep}{3pt}
\begin{tabular}{lllrrr}
\toprule
Encoder / data & Metric & Evaluation & Off & On & $\Delta$ [95\% CI]\\
\midrule
ImageNet / VisA & I-AUROC & Common & 80.79 & 73.49 & -7.30 [-11.90, -2.72]\\
ImageNet / VisA & I-AUROC & UCAD & 78.48 & 72.58 & -5.90 [-9.59, -1.98]\\
ImageNet / VisA & P-AP & UCAD & 23.61 & 18.89 & -4.72 [-9.89, -0.37]\\
DINOv3 / VisA & I-AUROC & Common & 82.56 & 85.67 & +3.12 [-0.47, +6.79]\\
DINOv3 / VisA & I-AUROC & UCAD & 82.58 & 87.55 & +4.96 [+2.21, +8.70]\\
DINOv3 / VisA & P-AP & Common & 38.86 & 41.21 & +2.35 [-5.42, +9.64]\\
DINOv3 / VisA & P-AP & UCAD & 29.94 & 28.86 & -1.08 [-7.84, +4.89]\\
DINOv3 / MVTec & I-AUROC & Common & 95.95 & 97.16 & +1.21 [+0.16, +2.70]\\
DINOv3 / MVTec & I-AUROC & UCAD & 96.22 & 97.20 & +0.98 [-0.10, +2.35]\\
DINOv3 / MVTec & P-AP & Common & 59.90 & 62.20 & +2.30 [-0.63, +5.69]\\
DINOv3 / MVTec & P-AP & UCAD & 57.07 & 59.05 & +1.98 [-0.76, +5.09]\\
ImageNet / MVTec & I-AUROC & Common & 91.07 & 91.59 & +0.52 [-1.25, +2.36]\\
ImageNet / MVTec & I-AUROC & UCAD & 90.26 & 90.45 & +0.19 [-1.39, +1.97]\\
ImageNet / MVTec & P-AP & Common & 52.02 & 51.92 & -0.10 [-2.15, +1.75]\\
ImageNet / MVTec & P-AP & UCAD & 43.47 & 43.37 & -0.10 [-2.51, +1.94]\\
\bottomrule
\end{tabular}

\end{table*}
\begin{figure*}[!htbp]
\centering\includegraphics[width=\textwidth]{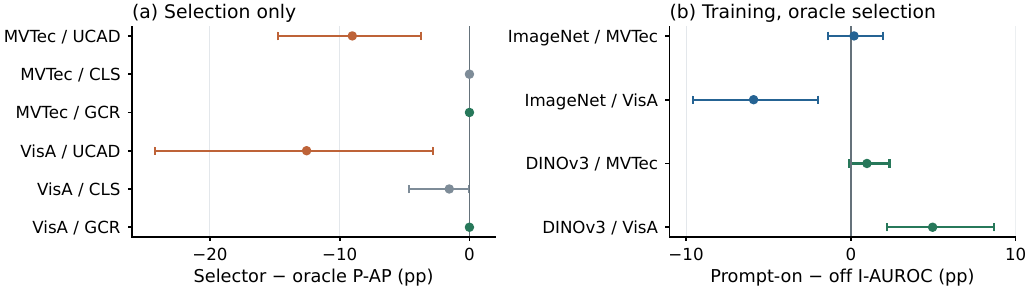}
\caption{\textbf{Separating routing loss from prompt-learning effects.} Left: prompt-free ImageNet \Pap{} difference between each selector and oracle routing. Right: prompted-minus-frozen \Iauc{} under oracle routing (ImageNet uses the public prefix here; see Table~\ref{tab:promptfix} for corrected retraining). Error bars indicate paired 95\% category-bootstrap intervals.}
\label{fig:selection}
\end{figure*}
The public ImageNet implementation reshapes the key/value prefix tensor without permuting its batch and key/value axes. At batch sizes $\ge2$, images can receive mismatched key-only or value-only prefixes. With the same \emph{Cable} prompt weights, for example, batch size 8 gives 71.31\% \Iauc{} / 19.39\% \Pap{}, while batch size 1 gives 45.99\% / 1.97\%. We correct the axis permutation, retrain prompts for 25 epochs, and compare them with frozen features under oracle routing (Table~\ref{tab:promptfix}).
\begin{table}[!htbp]
\caption{\textbf{Comparison of frozen features, original public prompts, and corrected retrained prompts on ImageNet.} Entries report \Iauc{} / \Pap{} (\%) under oracle selection with 196 references per bank and UCAD evaluation.}
\label{tab:promptfix}
\centering\small
\setlength{\tabcolsep}{2.5pt}
\begin{tabular}{lrrr}
\toprule
Dataset & Frozen & Original prompt & Corrected prompt\\
\midrule
MVTec & 90.26 / 43.47 & 90.45 / 43.37 & 90.16 / 42.12\\
VisA & 78.48 / 23.61 & 72.58 / 18.89 & 72.49 / 20.54\\
\bottomrule
\end{tabular}

\end{table}
The corrected ImageNet--VisA prompt remains 5.99 pp below the frozen detector in \Iauc{} ($-5.99$ pp [$-10.00$, $-2.23$]), with GCR matching oracle selection for both corrected conditions. DINOv3 uses a separate prefix adapter with the intended axis order. All prompt results use the final 25-epoch checkpoint.

\subsection{Comparing anomaly scores across products}
Category-mean AUROC ranks normal and defective images within each product. Task-pooled AUROC instead combines the three categories within each MVTec task ($5\times3$), adding cross-product pairs and changing their weighting on the same DNE predictions. Adapted DNE's advantage over Separate Min is 8.16 pp under pooled evaluation, compared with 1.37 pp under category means (main-paper Figure~\ref{fig:dne}). VisA has one category per task, so the two evaluations coincide.

\subsection{Consolidating category detectors into a shared model}
\begin{figure*}[!htbp]
\centering\includegraphics[width=\textwidth]{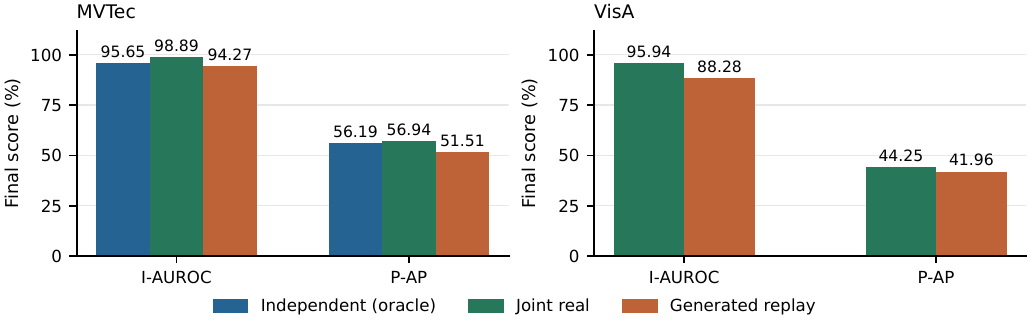}
\caption{\textbf{Comparing per-category real training, joint real training, and sequential generated replay in ReplayCAD.} \Iauc{} and \Pap{} under the official ReplayCAD evaluator (no independent per-category run is included on VisA).}
\label{fig:replay}
\end{figure*}

Independent real trains one ReplayCAD detector per category on real normal images and uses oracle selection. Joint real trains one detector on all real normal images together. Generated replay instead trains one detector sequentially with current real and synthetic historical images. Each uses the same architecture and evaluator and trains for 200 epochs on its respective inputs (Figure~\ref{fig:replay}).
On MVTec, Joint real replaces 15 separate detectors with one model and gains 3.24 pp \Iauc{} and 0.75 pp \Pap{} over Independent real. Generated replay instead finishes 1.38 pp and 4.68 pp below Independent real. Its gaps to Joint real are $-4.62$ pp [$-7.12$, $-2.40$] in \Iauc{} and $-5.43$ pp [$-9.08$, $-2.03$] in \Pap{} on MVTec, and $-7.66$ pp and $-2.29$ pp on VisA. Although all settings use 200 epochs, their dataset sizes and update counts differ. The replay generator is used during training and is excluded from deployed inference.

\section{What Does Preserving or Sharing a Detector Cost?}
\label{app:cost}
Preserving separate detectors and training a shared model require different amounts of storage and compute. We measure deployed state, acquisition compute, and end-to-end inference latency separately so that each cost has a defined boundary.

\subsection{Inference memory state vs.\ training-only assets}
\begin{table*}[!htbp]
\caption{\textbf{Deployed inference state and stream acquisition compute across 15 MVTec categories on one RTX 4090.} Inference state excludes training-only assets (such as generators or subspace bases). Counted acquisition operations are listed explicitly; full-stream wall-clock times including evaluation overhead are reported separately in Table~\ref{tab:streamtime}.}
\label{tab:acquisitioncost}
\centering\small
\setlength{\tabcolsep}{4pt}
\begin{tabular}{lrrl}
\toprule
System (15 MVTec categories) & State (GB) & GPU-h & Counted acquisition operations\\
\midrule
Frozen DINOv3 + GCR & 0.351884 & 0.097 & Feature extraction, coreset registration, registration checks\\
GCR + Dinomaly & 4.389779 & 5.443 & Expert training (5.346) and routing banks (0.097)\\
UCAD / ImageNet & 0.371693 & 1.759 & 25-epoch prompt training; masks and registration excluded\\
DNE $15\times1$ & 0.345610 & 0.633 & Training and Gaussian estimation\\
IUF $15\times1$ & 0.114500 & 19.084 & Task-process times, excluding 0.345 h of evaluation-only passes\\
ReplayCAD & 0.382802 & 50.130 & Generator (45.76), sampling (1.06), detector (3.31)\\
CDAD $15\times1$ & 4.398400 & 59.990 & Training (38.22) and basis construction (21.77)\\
\bottomrule
\end{tabular}

\end{table*}
\begin{figure*}[!htbp]
\centering\includegraphics[width=\textwidth]{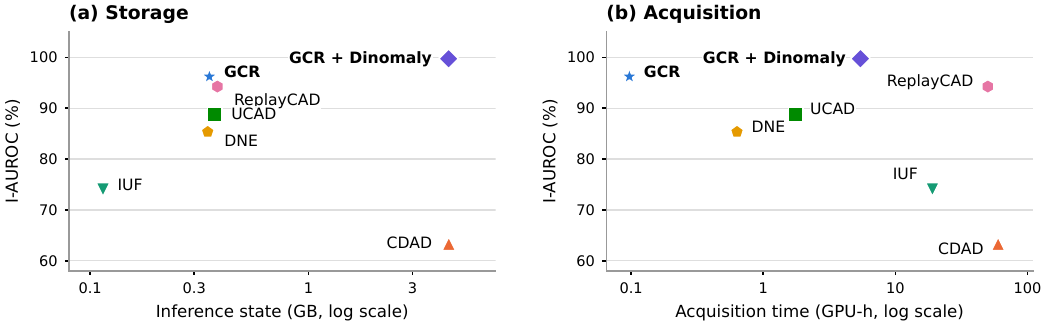}
\caption{\textbf{Image-level detection (\Iauc{}) vs.\ deployed storage and acquisition time.} Companion to main-paper Figure~\ref{fig:resources}, including image-level DNE ($15\times1$ stream). Accounting boundaries follow Table~\ref{tab:acquisitioncost}.}
\label{fig:resourceimage}
\end{figure*}
\begin{figure*}[!htbp]
\centering
\includegraphics[width=\textwidth]{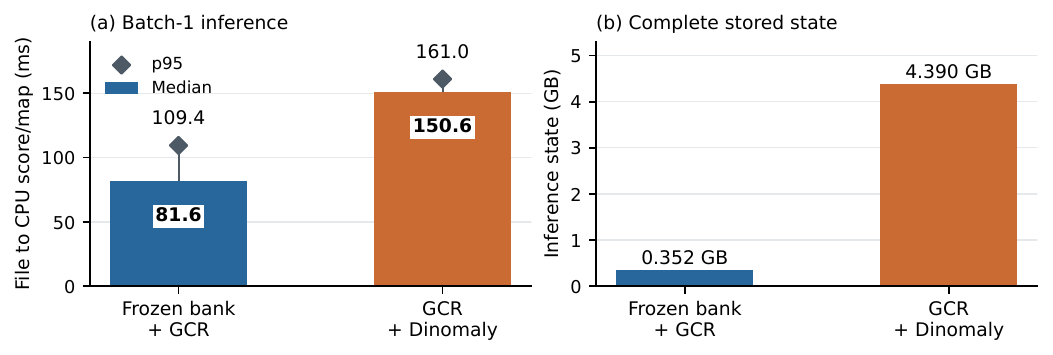}
\caption{\textbf{End-to-end inference latency and deployed state for preserved detectors.} Left: batch-1 FP32 latency from file read to CPU score/map on an RTX 4090 (233 images, 3 passes after 10 warm-up images; diamonds mark p95 latency). Right: total deployed inference state across 15 MVTec categories (with shared backbone weights counted once).}
\label{fig:deployment}
\end{figure*}
In decimal units ($10^6$ bytes/MB and $10^9$ bytes/GB), frozen DINOv3 with GCR occupies 351.884 MB: a 342.852 MB backbone and 9.032 MB for 15 reference banks, each containing 602,112 bytes of FP32 features (Table~\ref{tab:acquisitioncost}, Figure~\ref{fig:deployment}). Preserved Dinomaly detectors with GCR banks occupy 4,389.779 MB when shared encoder weights are counted once, adding about 246 MB per category.

The backbone accounts for most of the frozen-bank system's storage. Shrinking all 15 banks from 196 to 13 references saves 8.43 MB but sharply lowers localization (Section~\ref{app:budget}). A 115 MB total budget would therefore require a smaller or compressed backbone as well as smaller banks.
\begin{table*}[!htbp]
\caption{\textbf{Breakdown of deployed inference state vs.\ training-only checkpoints and auxiliary assets.}}
\label{tab:storageparts}
\centering\small
\begin{tabular}{lr}
\toprule
Stored component & Size\\
\midrule
DNE inference state & 345.6 MB\\
DNE next-stage checkpoint & 1.07 GB\\
IUF final model & 114.5 MB\\
ReplayCAD detector & 382.8 MB\\
ReplayCAD personalized generator checkpoint & 0.8 MB\\
ReplayCAD shared latent diffusion model & 6.15 GB\\
CDAD final model & 4.40 GB\\
CDAD same-task resume checkpoint & 11.13 GB\\
CDAD next-task basis & 1.75 GB\\
\bottomrule
\end{tabular}

\end{table*}
Table~\ref{tab:storageparts} separates deployed state from files used to continue training, including optimizer checkpoints, generators, and projection bases. UCAD's approximately 372 MB includes its shared ViT-B backbone, prompts, keys, and knowledge banks. Runtime GPU memory is measured separately: batch-1 inference peaks at 4.7 GB for frozen banks (12.6 images/s) and 4.5 GB for preserved Dinomaly (7.3 images/s).

\subsection{Acquisition time and training workloads}
\begin{table*}[!htbp]
\caption{\textbf{Breakdown of detector training and registration times.} UCAD times reflect 25-epoch prompt training across summed workers, excluding SAM mask generation and bank registration.}
\label{tab:trainingcost}
\centering\small
\begin{tabular}{llr}
\toprule
Configuration & Workload & Time\\
\midrule
UCAD ImageNet prompts & 15 MVTec categories, 25 epochs, summed workers & 6,330.8 s\\
UCAD DINOv3 prompts & 15 MVTec categories, 25 epochs, summed workers & 1,870.9 s\\
DNE Adapted & MVTec $5\times3$, per order & 1,942--2,000 s\\
DNE three frozen conditions & MVTec $5\times3$, summed conditions per order & 40--48 s\\
DNE Adapted & VisA $12\times1$, per order & 4,722--4,979 s\\
DNE three frozen conditions & VisA $12\times1$, summed conditions per order & 104--113 s\\
Frozen DINOv3 + GCR & 15 MVTec banks, 196 references per category & 0.097 GPU-h\\
ReplayCAD Independent & 15 independent MVTec detector trainings & 3.88 GPU-h\\
ReplayCAD Joint real & One real-data MVTec detector & 2.24 GPU-h\\
ReplayCAD generator training & MVTec official rerun & 45.76 GPU-h\\
ReplayCAD image generation & MVTec official rerun & 1.06 GPU-h\\
ReplayCAD detector training & MVTec official rerun & 3.31 GPU-h\\
ReplayCAD official rerun & Generator training + sampling + detector training & 50.13 GPU-h\\
\bottomrule
\end{tabular}

\end{table*}
Table~\ref{tab:trainingcost} separates the acquisition workload into its main operations. ReplayCAD uses 50.13 GPU-h on MVTec: 45.76 for generator training, 1.06 for image sampling, and 3.31 for detector training. Table~\ref{tab:streamtime} instead reports elapsed time for the complete single-category streams, including intermediate evaluation passes.
\begin{table*}[!htbp]
\caption{\textbf{Total wall-clock time (including per-task evaluation overhead) for the full streams in Table~\ref{tab:fullstreamsvariants}.}}
\label{tab:streamtime}
\centering\small
\setlength{\tabcolsep}{3pt}
\begin{tabular}{llr}
\toprule
Dataset & Full-stream implementation & Time (h)\\
\midrule
MVTec & CDAD / released code & 66.6\\
MVTec & CDAD / paper Eq.~(6) & 55.6\\
MVTec & IUF / released code & 19.5\\
VisA & CDAD / released code & 150.7\\
VisA & CDAD / paper Eq.~(6) & 139.5\\
VisA & IUF / released code & 68.8\\
\bottomrule
\end{tabular}

\end{table*}
In the two-stage \emph{Bottle} $\rightarrow$ \emph{Cable} transition (200 epochs), onboarding \emph{Cable} takes 1 h 01 m 0.8 s of prompt training plus 19.0 s of bank registration in UCAD; 1 h 47 m 50.2 s in ReplayCAD (59 m 38.3 s detector training, 37 m 21.5 s generator training, and 10 m 50.3 s to sample 800 images); and 2 h 27 m 40.1 s in CDAD when reusing the \emph{Bottle} subspace basis.

\subsection{End-to-end inference latency}
Given $|\mathcal C_t|$ candidate categories, $M_r$ sampled test patches, $K$ references per bank, and feature dimension $d$, GCR routing requires $O(|\mathcal C_t|M_rKd)$ operations and stores $Kd$ floats per bank. Table~\ref{tab:resources} reports end-to-end inference latency covering both routing and downstream anomaly map generation.

\begin{table*}[!htbp]
\caption{\textbf{End-to-end inference latency across deployment configurations (RTX 4090, FP32, batch size 1).} The top group evaluates full 15-category MVTec deployments; the middle group evaluates two-stage (\emph{Bottle}--\emph{Cable}) checkpoints at each method's native resolution; the bottom row evaluates image-score-only DNE.}
\label{tab:resources}
\centering\small
\setlength{\tabcolsep}{5pt}
\begin{tabularx}{\textwidth}{lrrX}
\toprule
System & Median (ms) & p95 (ms) & Measured checkpoint and output boundary\\
\midrule
\multicolumn{4}{l}{\emph{Matched 15-category deployment; 233 images, three passes}}\\
Frozen DINOv3 + GCR & 81.6 & 109.4 & All 15 banks; file reading to CPU score and map\\
GCR + Dinomaly & 150.6 & 161.0 & All 15 experts resident; file reading to CPU score and map\\
\midrule
\multicolumn{4}{l}{\emph{Bottle--Cable two-stage checkpoints; 233 images, three passes}}\\
UCAD ImageNet & 81.7 & 91.5 & Native resolution; file reading to CPU score and map\\
ReplayCAD & 85.6 & 104.2 & Native resolution; file reading to CPU score and map\\
IUF & 128.4 & 174.0 & Native resolution; file reading to CPU score and map\\
CDAD & 944.3 & 1414.9 & Native resolution; file reading to CPU score and map\\
\midrule
\multicolumn{4}{l}{\emph{Separate image-score-only measurement; 128-image subset}}\\
DNE & 65.3 & 106.5 & Final state; CPU image score, no pixel map\\
\bottomrule
\end{tabularx}

\end{table*}
For the full 15-category deployments, we synchronize CUDA and time from disk file reading through GCR routing and anomaly scoring to CPU-returned scores and maps. Three passes use the same 233 \emph{Bottle} and \emph{Cable} images after 10 warm-up images, with all 15 detectors resident.

The two-stage checkpoints retain native input resolutions: 224 for UCAD, 256 for ReplayCAD and CDAD, and 448 for IUF. Their median latencies are 81.68, 85.57, 944.32, and 128.44 ms, respectively. A separate GCR--CLS comparison on the frozen bank system gives 83.45 ms and 83.30 ms, respectively. DNE's 65.3 ms measurement uses 128 images and returns only an image score.

\FloatBarrier
\end{document}